\documentclass[lettersize,journal]{IEEEtran}
\usepackage{amsmath,amsfonts}
\usepackage{array}
\usepackage[caption=false,font=normalsize,labelfont=sf,textfont=sf]{subfig}
\usepackage{textcomp}
\usepackage{stfloats}
\usepackage{url}
\usepackage{float}
\usepackage{verbatim}
\usepackage{graphicx}
\usepackage{cite}
\usepackage{afterpage}
\usepackage{bbm}
\usepackage{mathrsfs}
\usepackage{hyperref}
\hypersetup{
    colorlinks=true,
    linkcolor=blue,
    citecolor=blue,
    urlcolor=blue
}
\usepackage{wrapfig}
\usepackage{orcidlink}
\newtheorem{definition}{\textbf{Definition}}
\newtheorem{problem}{\textbf{Problem}}

\makeatletter
\newcommand{\removelatexerror}{\let\@latex@error\@gobble}
\makeatother
\usepackage[ruled,linesnumbered]{algorithm2e}
\usepackage{amsmath}
\usepackage{amssymb}

\usepackage{xcolor}
\usepackage{etoolbox}
\usepackage[numbers,sort&compress]{natbib}
\usepackage[font=small,labelfont=bf,margin=\parindent,tableposition=top]{caption}
\usepackage{subcaption}
\usepackage{titlesec}

\titlespacing*{\section}{0pt}{0.5\baselineskip}{0.2\baselineskip}
\titlespacing*{\subsection}{0pt}{0.35\baselineskip}{0.35\baselineskip}

\usepackage{float}

\makeatletter
\pretocmd{\NAT@open}{\begingroup\color{blue}}{}{}
\apptocmd{\NAT@close}{\endgroup}{}{}
\makeatother
\begin{document}

\title{Deep Reinforcement Learning Enabled Persistent Surveillance with Energy-Aware UAV-UGV Systems for Disaster Management Applications}

\author{Md Safwan Mondal$^{\orcidlink{0000-0003-4757-9403}}$, \IEEEmembership{ Student Member, IEEE}, Subramanian Ramasamy$^{\orcidlink{0000-0002-0665-3243}}$,   Pranav Bhounsule$^{\orcidlink{0000-0002-7504-6009} }, \ \IEEEmembership{Member, IEEE}$
\thanks{Md Safwan Mondal, Subramanian Ramasamy and 
Pranav A. Bhounsule are with the Department of Mechanical
and Industrial Engineering, University of Illinois Chicago, IL,
60607 USA.
        {\tt\small mmonda4@uic.edu}, 
        {\tt\small sramas21@uic.edu},   
        {\tt\small pranav@uic.edu}
   }%
   \thanks{ $\dagger$ Corresponding author, *This work was supported by ARO grant
W911NF-14-S-003. }
}

\markboth{Journal of \LaTeX\ Class Files,~Vol.~14, No.~8, August~2021}%
{Shell \MakeLowercase{\textit{et al.}}: A Sample Article Using IEEEtran.cls for IEEE Journals}


\maketitle

\begin{abstract}
Integrating Unmanned Aerial Vehicles (UAVs) with Unmanned Ground Vehicles (UGVs) provides an effective solution for persistent surveillance in disaster management. UAVs excel at covering large areas rapidly, but their range is limited by battery capacity. UGVs, though slower, can carry larger batteries for extended missions. By using UGVs as mobile recharging stations, UAVs can extend mission duration through periodic refueling, leveraging the complementary strengths of both systems. To optimize this energy-aware UAV-UGV cooperative routing problem, we propose a planning framework that determines optimal routes and recharging points between a UAV and a UGV. Our solution employs a deep reinforcement learning (DRL) framework built on an encoder-decoder transformer architecture with multi-head attention mechanisms. This architecture enables the model to sequentially select actions for visiting mission points and coordinating recharging rendezvous between the UAV and UGV. The DRL model is trained to minimize the \textit{age periods} (the time gap between consecutive visits) of mission points, ensuring effective surveillance. We evaluate the framework across various problem sizes and distributions, comparing its performance against heuristic methods and an existing learning-based model. Results show that our approach consistently outperforms these baselines in both solution quality and runtime. Additionally, we demonstrate the DRL policy's applicability in a real-world disaster scenario as a case study and explore its potential for online mission planning to handle dynamic changes. Adapting the DRL policy for priority-driven surveillance highlights the model's generalizability for real-time disaster response. More details are available on our website: \href{https://sites.google.com/view/ecucps}{ https://sites.google.com/view/ecucps}.

\end{abstract}


\section{Introduction}

\IEEEPARstart{U}{nmanned} Aerial Vehicles (UAVs) have emerged as transformative agents in modern disaster management systems due to their speed, mobility, cost-efficiency, and ability to navigate challenging environments \cite{rajan2021disaster, erdelj2017help}. Amid the increasing frequency of natural disasters worldwide, the need for rapid and effective disaster response is crucial to save lives and prevent substantial economic and infrastructure losses. This necessitates continuous monitoring and data collection at disaster sites, where UAV-based intelligence, surveillance, and reconnaissance (ISR) missions have become integral to contemporary disaster management strategies \cite{chaturvedi2019comparative, raffetto2004unmanned}. UAVs can provide vital real-time data and continuously assess scenarios to support both the planning and operational phases of disaster response. For instance, during the 2015 Nepal earthquake, UAVs were used for rapid damage assessment in areas inaccessible to rescue teams, providing high-resolution images that helped to determine the scale of devastation and identify safe paths for ground teams \cite{yamazaki2015construction}. Similarly, during the severe floods in Kerala in 2018, drones were deployed to locate missing people and assess the damage \cite{salmoral2020guidelines}. During the COVID-19 pandemic, drones assisted Indian officials in curbing the virus's spread by conducting street surveillance to monitor illegal gatherings, measuring body temperatures, and facilitating the delivery of medical supplies to isolated zones in cities like Delhi and Bengaluru \cite{gupta2021uses}.

\begin{figure}[t]
\centering
\includegraphics[scale=0.525
]{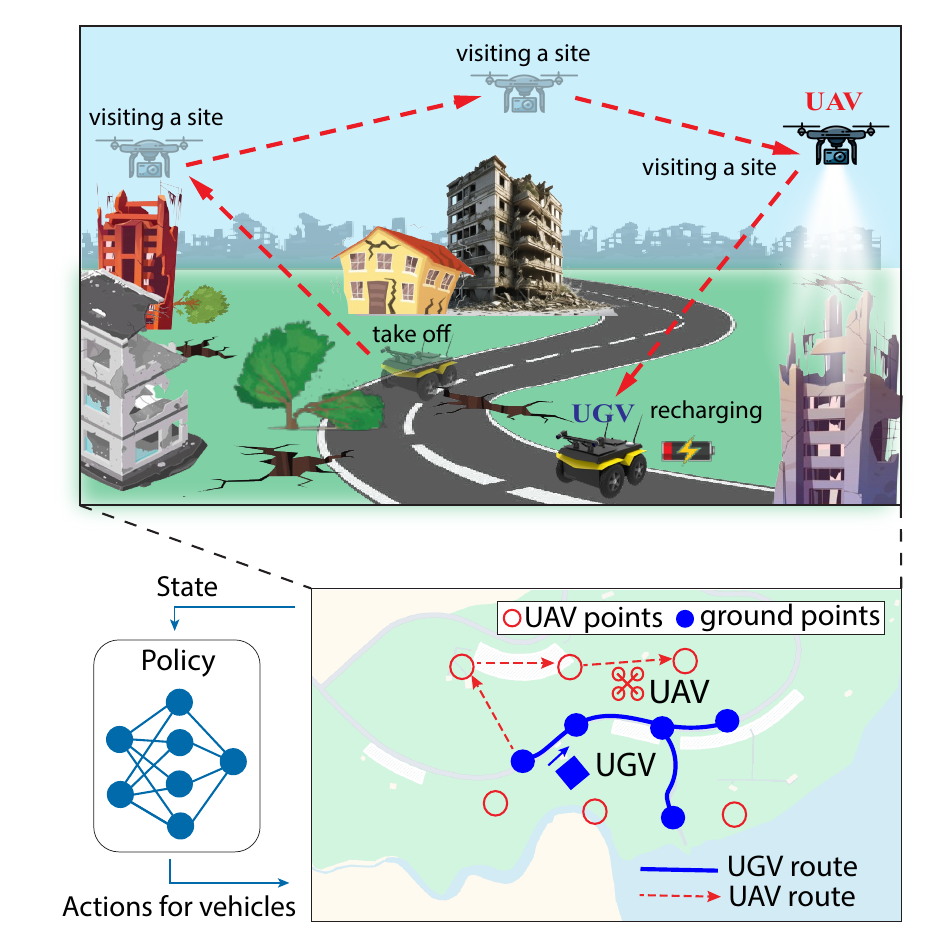}
\caption{Illustration of collaboration between an energy-constrained UAV and a UGV for surveying disaster-stricken areas. The UAV performs continuous surveillance and recharges through the UGV. The proposed DRL policy determines mission point visits and coordinates UAV-UGV recharging.}
\label{problem}
\end{figure}

Despite their advantages, UAVs face significant limitations due to restricted battery life, which can hinder continuous operation during prolonged missions. To address this, UAVs can be periodically recharged at fixed depots; however, the stationary nature of these refueling points can limit their operational range. Consequently, Unmanned Ground Vehicles (UGVs) are employed as mobile recharging bases, providing essential ground support and extending the operational capabilities of UAVs along designated roadways \cite{lin2022robust, yu2019coverage, maini2015cooperation}. This synergy between UAVs and UGVs enhances the efficiency and effectiveness of persistent surveillance and comprehensive disaster management coverage.

Our paper explores a UAV-based persistent surveillance problem integrated with a UGV to overcome endurance limitation and maximize operational impact in disaster scenarios. Specifically, in a given scenario, an energy-constrained UAV starts its journey from a designated depot to visit a set of mission points and collect information (see Figure \ref{problem})  For sustained operations, the UAV is periodically recharged by a UGV, which also departs from the same depot and travels along the road network to visit and collect information from mission points located on the road. The objective is to minimize the time gap between consecutive visits to mission points by maximizing the visit frequency of either vehicle. This energy-constrained UAV-UGV cooperative persistent surveillance (\textbf{ECUCPS}) system is crucial for disaster management scenarios, where continuous and up-to-date information collection from disaster sites is essential. The integration of UAV with UGV ensures that critical mission points are frequently visited, providing reliable real-time data to support effective disaster response efforts.

In the literature, the cooperative routing problem has been analyzed as a variant of the traveling salesman problem (TSP) or the vehicle routing problem (VRP) \citep{wang2019vehicle, tang2019study}. Mixed integer linear programming (MILP) formulations have been proposed to obtain exact solutions to different variants of this cooperative routing problem \cite{sundar2016formulations, murray2015flying}. However, being an NP-hard combinatorial optimization problem, it becomes intractable as the size of the problem scenarios grows \cite{cattaruzza2016vehicle}. As a result, heuristics such as genetic algorithms, tabu search, and ant colony optimization are utilized to produce high-quality approximate solutions for these optimization problems within acceptable time frames \cite{ropero2019terra, chen2019path, zhang2022cooperative}. These heuristics, however, are typically tailored to specific problem types, making the process labor-intensive and limiting their generalizability and effectiveness. 

In recent years, learning-based methodologies have emerged as a promising alternative for solving the vehicle routing problem, its variants, and other combinatorial optimization problems \cite{kaempfer2018learning, wang2020learning, sykora2020multi, paul2022learning}. Deep reinforcement learning (DRL)-based frameworks have been employed to address the TSP and VRP due to their ability to autonomously discover optimal routing strategies, handle complex constraints, and adapt to diverse problem scenarios through continuous learning and improvement. While DRL has successfully addressed a variety of vehicle routing problems, the UAV-UGV collaborative routing problem, particularly in the context of persistent surveillance, remains relatively unexplored. Unlike traditional routing tasks, where previously visited points are excluded from the action space, persistent surveillance requires continuous revisitation of mission points, creating a constant action space and adding layers of complexity. Also, incorporating ground vehicles as mobile
recharging stations to increase the operational range of UAVs expands the potential recharging locations and leads to a much larger and more dynamic solution space compared to
the relatively static and predictable nature of fixed recharging stations. Consequently, the UAV-UGV cooperative surveillance problem becomes computationally more challenging and requires advanced optimization techniques to ensure both efficiency and effectiveness. Additionally, in cooperative routing, it is essential not only to determine the optimal sequence of mission point visits but also to strategically plan the recharging instances between UAVs and UGVs to ensure seamless coordination. Ramasamy et al. \cite{ramasamy2022heterogenous} demonstrated the impact of these recharging instances on the overall efficiency of cooperative routing, underscoring the need for effective synchronization between vehicles. Building on our prior work \cite{mondalattention} on DRL-based cooperative routing framed as a TSP problem, we extend that approach to develop a more advanced framework to address the specific challenges of persistent surveillance. This paper proposes a deep reinforcement learning (DRL) framework based on an encoder-decoder transformer architecture and policy gradient method to determine coordinated routing between an energy-constrained UAV and a UGV acting as a mobile recharging station. The framework has been tested across different problem sizes and distributions, indicating its generalizability. To this end, we present the following novel contributions:

1. \textbf{Modeling the Problem:} We first model the ECUCPS problem using a multi-level optimization strategy for approximate heuristic solutions and then as a Markov Decision Process (MDP) for solving it through reinforcement learning (RL).

2. \textbf{Framework Evaluation:} We introduce an encoder-decoder-based transformer architecture with attention layers and dynamic node embeddings to tackle the ECUCPS problem. The framework's generalization capabilities are demonstrated through systematic evaluations across various problem sizes and distributions, and its applicability is further validated in a simulation case study of a real-world disaster scenario.

3. \textbf{Comparison with Existing Methods:} We compare our method against heuristic-based approaches for persistent surveillance routing and a learning-based framework as baselines. Our approach demonstrates superior solution quality, highlighting its effectiveness and reliability over the baselines.

4. \textbf{Online and Priority-Driven Route Planning:} We investigate the framework's utility for online route planning in response to dynamically appearing mission points and also extend its application for prioritized visits to critical mission points in weighted persistent surveillance scenarios.

The remainder of this paper is structured as follows: Section \ref{sec2} reviews related work. Section \ref{sec3} presents the problem statement and Markov Decision Process (MDP) formulation. Section \ref{sec4} describes problem modeling and a heuristics approach for persistent surveillance. Section \ref{sec5} details the proposed method and policy network. Section \ref{sec6} presents and analyzes our experimental results. Finally, Section \ref{sec7} concludes the paper and discusses future work.

\begin{figure*}[htbp]
\centering
\includegraphics[scale=0.475
]{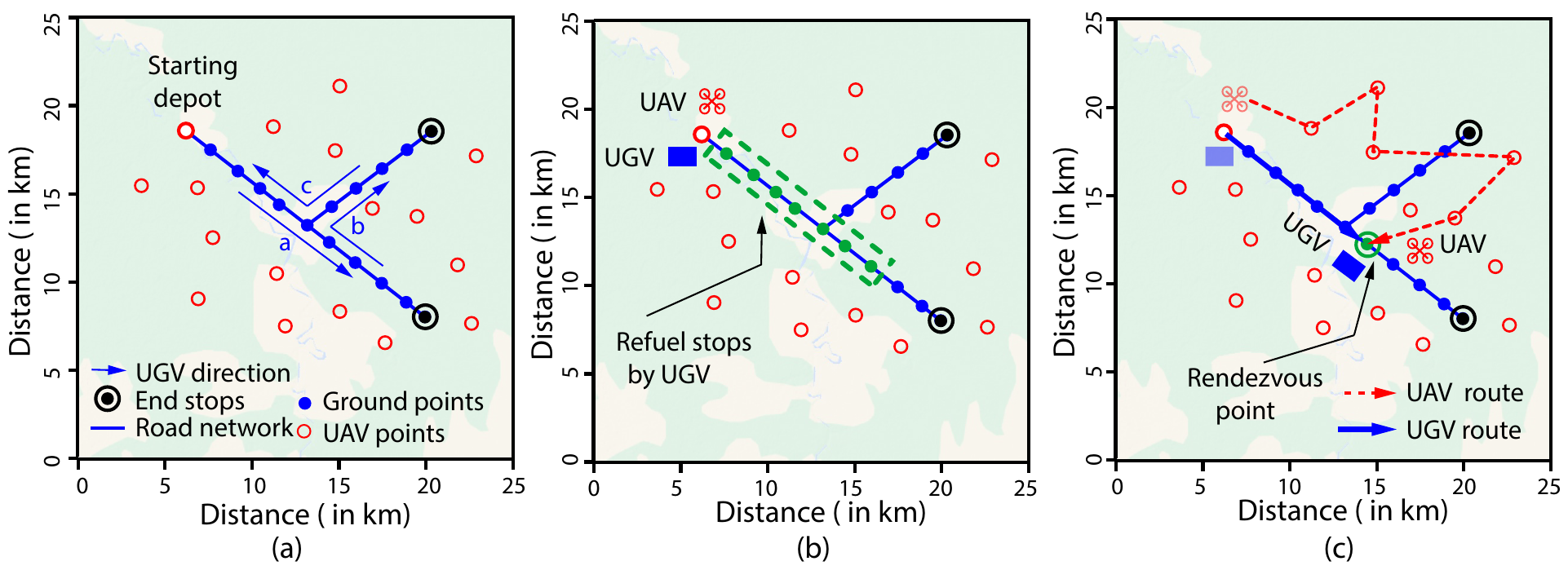}
\caption{Bilevel optimization workflow: a) Given scenario with UAV and ground points, and UGV's traversal direction along the road network from the starting depot as obtained from the TSP solution. b) Available refuel stops provided by the UGV during O-EVRPTW route planning for the UAV. c) Recharging rendezvous between the UAV and UGV, along with their respective route sorties, derived from the O-EVRPTW solution.}
\label{UGVPlanner}
\end{figure*}

\section{Related works}
\label{sec2}
In this section, we review the literature in two key areas: UAV-based routing for persistent surveillance and recent advancements in learning-based approaches for routing problems.

\subsection{UAV based Persistent Surveillance Problem}
In recent years, UAV routing problems have garnered significant attention in modern disaster management due to their potential to address complex logistical challenges. Specifically, models such as the Traveling Salesman Problem with Drones (TSP-D) and the Vehicle Routing Problem with Drones (VRP-D) have been pivotal in developing efficient solutions for these challenges. Scherer et al. \cite{scherer2016persistent} explored the energy and communication constraints of UAVs, developing an offline path planning algorithm for multiple UAVs and assessing the impact of base station configurations on mission performance. Concurrently, Angun and Dundar \cite{angun2020intelligent} approached the surveillance problem from an energy-capacitation perspective, framing it as a location routing problem suitable for the risk mitigation phase of disaster management. In the context of emergency operations, Calamoneri et al. \cite{calamoneri2024management} proposed the Multi-Depot Multi-Trip Vehicle Routing Problem with Total Completion Times Minimization (MDMT-VRP-TCT). Their model facilitates a fleet of UAVs in visiting disaster-affected sites multiple times from various depots, thereby enhancing rescue efforts.
Nigam et al. \cite{nigam2014multiple} addressed the persistent surveillance problem by segmenting a specified 2D area into a grid map tailored to the sensor footprint, assigning UAVs to repeatedly monitor these grids. Further innovations in this field include the work of Stump and Michael \cite{michael2011persistent}, who employed the Vehicle Routing Problem with Time Windows, utilizing a receding horizon strategy to maintain ongoing surveillance over discrete targets. Despite these advancements, the issue of UAV fuel limitations has been relatively underexplored. Hari et al. \cite{hari2020optimal} introduced a model that restricts the number of UAV visits to reflect fuel constraints, determining the minimal maximum revisit duration among target nodes. Traditional setups assume the existence of fixed service stations for UAV recharging. However, recent studies suggest employing mobile UGVs as dynamic recharging units to enhance operational efficiency, an approach known as the UAV-UGV cooperative routing problem. This integration of heterogeneous vehicles complicates the problem, making exact Mixed Integer Linear Programming (MILP) solutions impractical due to scalability issues. To tackle this, heuristic methods have been devised to achieve high-quality, feasible solutions within reasonable time frames. For instance, Seyedi et al. \cite{seyedi2019persistent, lin2022robust} developed a heuristic that organizes a uniform UAV-UGV team to cyclically patrol optimally partitioned surveillance areas. While effective for area coverage, this strategy may not suit scenarios with discrete mission points due to its reliance on spatial partitioning. In heterogeneous vehicle cooperative routing, a prevalent tactic is the multi-level or multi-echelon optimization approach, such as ``UGV first, UAV second," which simplifies the complexity by breaking down the problem into more manageable subproblems. Maini et al. \cite{maini2015cooperation, maini2019cooperative} implemented a bi-level strategy, initially identifying potential rendezvous points via a minimum set cover problem, followed by formulating a MILP to ascertain optimal routes for both UAV and UGV. Innovatively, Nigam et al. \cite{nigam2008persistent} proposed a value-sum strategy directing UAVs towards areas where the cumulative value is highest or a target-based tactic focusing on cells with the greatest singular value determined by latency periods. Similarly, Chour et al. \cite{chour2022reactive} tackled a UAV-UGV rendezvous problem by creating a multi-level coordination, scheduling, and planning algorithm. They solved a reward-based traveling salesman problem using a receding horizon approach to establish the UAV’s route for UGV-based recharging. However, this method bears the inherent limitation of the \textit{horizon effect} \cite{asghar2023risk}, where a route deemed optimal within a planning horizon may prove suboptimal over the complete mission duration.

\subsection{DRL for UAV-UGV cooperative routing}

In recent times, learning-based methods have emerged as a viable alternative for addressing the routing problem, its variants, and other combinatorial optimization challenges. Most deep reinforcement learning (DRL) approaches involve end-to-end encoding and decoding processes for solving routing problems. The encoder extracts node features from inputs, while the decoder utilizes these features to sequentially select node sequences, training the neural model through reinforcement learning to preferentially output high-profit nodes at each time step. Vinyals et al. \cite{vinyals2015pointer} pioneered the Pointer Network (PN), enhancing a recurrent neural network (RNN) with an attention mechanism \cite{bahdanau2014neural} and training it in a supervised manner to predict optimal tours for the Traveling Salesman Problem (TSP). However, supervised learning methods for routing are constrained by the need for extensive and costly labels generated by pre-solving the routing problems which is a time-consuming process. Addressing these limitations, Kool et al. \cite{kool2018attention} introduced a Transformer-based encoder-decoder architecture that outperformed various traditional heuristic methods across a spectrum of routing problems. Similarly, Li et al. \cite{li2021deep} utilized a DRL strategy with attention mechanisms to significantly enhance solution quality and computational efficiency in tackling the heterogeneous capacitated vehicle routing problem. Furthermore, Wu et al. \cite{wu2021reinforcement} explored the truck-and-drone-based last-mile delivery problem using reinforcement learning. They effectively divided the optimization task into customer clustering and routing stages, employing an encoder-decoder framework with RL to address the challenge. Similarly, Fan et al. \cite{fan2022deep} employed a multi-head attention mechanism alongside a DRL policy to design routes for an energy-constrained UAV, but their model relied on fixed UAV recharging stations. Bana et al. \cite{bana2024deep} also applied DRL to persistent surveillance for fuel-limited UAVs, though their approach was similarly constrained by the assumption of fixed recharging stations. Building on these insights, this paper introduces a novel DRL framework that leverages an encoder-decoder transformer architecture and a policy gradient method to optimize coordinated routing between an energy-limited UAV and a UGV acting as a mobile recharging station. Tested across diverse problem sizes and scenarios, the proposed framework demonstrates significant adaptability and generalizability, offering a promising solution for UAV-UGV cooperative routing for persistent surveillance.

\section{Problem overview}
\label{sec3}
\subsection{Problem statement}
As the exemplary application of UAV-UGV collaborative routing for disaster management, we focus on a persistent surveillance problem. Persistent surveillance involves monitoring an area or site for an extended period to collect real-time information continuously. This approach plays a crucial role in both pre- and post-disaster scenarios, assisting in tasks such as damage assessment, search and rescue operations, delivery of emergency relief supplies, and other critical functions. To define the problem formally, let us consider a system consisting of a UAV $U_a$ and a UGV $U_g$ tasked with visiting a set of $n$ mission points $\mathcal{M} = \{m_1, m_2, ..., m_n\}$ spread over a region (see Figure \ref{UGVPlanner}\textcolor{blue}{a}). There are two types of mission points: those located within a road network 
$G$, which can be surveyed either by UAV flyover or by UGV-based road visit (referred to as ground
points, $\mathcal{M}_g$), and those situated outside the road network, accessible only to the UAV (referred to as UAV points, $\mathcal{M}_a$). Thus, the set of mission points $\mathcal{M}$ can be defined as $\mathcal{M}$ = $\mathcal{M}_g \cup \mathcal{M}_a$.
 
\begin{definition}
The \textit{age period} of any mission point is defined as the time gap between the current time and the last time that particular mission point was visited.  
\begin{equation}
\mathfrak{a}^k  = t - t^k_{\text{last}}
\end{equation}
here, $\mathfrak{a}^k$ is the \textit{ age period} for mission point $m_k \in \mathcal{M}$ at time $t$ and $t^k_{last}$ is the last time when the mission point was visited.

\end{definition}

\begin{definition}
To reinforce persistent visits to the mission points and ensure that the most recent data are always collected from the disaster sites, we define the score metric as:
\begin{equation}
\mathbb{S} = \frac{1}{\alpha}\sum_{k=1}^n \sum_{q=1}^p \left(t_q^k-t_{q-1}^k\right)^2 \ = \frac{1}{\alpha} \sum_{k=1}^n \sum_{q=1}^p (\mathfrak{a}^k_q)^2  
\end{equation}
Here, $t^k_0, t^k_1,...,t^k_p$ are the time instances in the ascending order when a mission point $m_k$ is visited during the mission period. The parameter $\alpha = T_m^2$  is used to scale down the score metric value where $T_m$ is the mission period duration. The score metric is designed as a square function of the \textit{age period} to penalize longer intervals between consecutive visits to a mission point.
\end{definition}
The UAV and UGV share heterogeneous characteristics: the UAV has a limited battery capacity $F^a$ but can fly at a faster velocity 
$v_a$, whereas the UGV moves at a slower pace 
$v_g$ and operates exclusively on the road network 
$G$. The UGV can refuel the UAV by acting as a mobile recharging station. For recharging, the UAV meets the UGV at any ground point, spends a fixed time recharging $T_R$ there, and then resumes its flight by taking off from the UGV. The UAV and UGV operate continuously to visit mission points and meet for recharging rendezvous till the end of mission period duration.

\begin{problem}
The objective is to devise a strategy for collaborative operation between the UAV and the UGV to persistently visit all mission points in the scenario during the mission period, thereby minimizing the score metric at the end of the mission planning horizon. This challenge is multifaceted, requiring the optimization of the UAV and UGV route sorties to visit mission points and the scheduling of their recharging instances to synchronize them in both time and location.
\end{problem}

\subsection{Markov Decision Process (MDP) formulation}

\begin{figure}[htbp]
\vspace{-2mm}
\centering
\includegraphics[scale=0.45
]{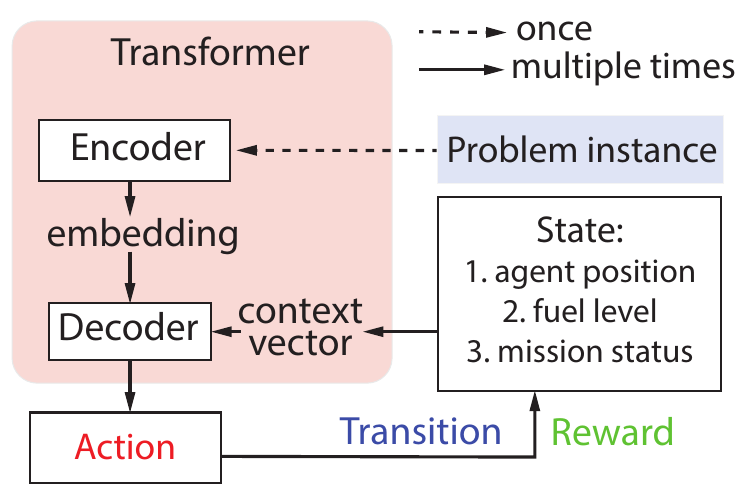}
\caption{MDP representation for the UAV-UGV cooperative persistent surveillance problem utilizing a Transformer architecture.}
\label{MDP}
\vspace{1mm}
\end{figure}
The problem can be modeled as a sequential decision-making system where the agent sequentially selects the mission points to visit (see Figure \ref{MDP}). This system can be formulated as a Markov Decision Process (MDP), with its components defined by the tuple $ < \mathcal{S}, \mathcal{A}, \mathcal{R}, \mathcal{T}> $, as follows: \\ 
1) \textbf{State Space\ $(\mathcal{S})$:} At any decision making step, the state of the environment $s_t \in \mathcal{S}$ is defined as, $ s_t = (p_t, f_t, \ q_t) $. Here, $ p_t = \{x_t, y_t\}$ represents the current position of the agent and $f_t$ indicates its fuel level. Additionally, $q_t = \{x^i, y^i, \mathfrak{a}^i_t\}$ highlights the coordinates and \textit{age period} status of the mission points $m_i \in \mathcal{M} $. \\
2) \textbf{Action Space\ $(\mathcal{A})$:} The selection of a mission point is defined as the action $a_t \in \mathcal{A}$. The agent can perform two types of actions: \textit{visiting} (on both UAV points $\mathcal{M}_a$ and ground points $\mathcal{M}_g$) or \textit{recharging} (only on ground points $\mathcal{M}_g$). Since $\mathcal{M}_g$ is used for both visiting and recharging, the action space is defined as, $\mathcal{A} = \{\mathcal{M}_g (\text{for \textit{recharging}}) + \mathcal{M}_g (\text{for \textit{visiting}}) + \mathcal{M}_a (\text{for \textit{visiting}}) \}$. Infeasible actions based on the current state $s_t$ at time $t$ are masked out from the action space (details in subsection \ref{decoder_sub}).  \\
3) \textbf{Reward\ (\(\mathcal{R}\)):} To align with the objective of minimizing the total score metric, the reward \(\mathcal{R}\) is defined as the negative of this metric. Specifically, at each decision-making step \(t\), the reward \(r_t\) is calculated as the square of the \textit{age period}, \(r_t = (\mathfrak{a}^i_t)^2\), where \(\mathfrak{a}^i_t\) represents the \textit{age period} of the mission point \(m_i\) selected as the action \(a_t\) at step \(t\). The total reward for the mission is computed as the negative sum of all such rewards over the entire mission planning horizon: $\mathcal{R} = - \sum_{t=0}^{T_m} r_t $. Finally at the end of the mission planning horizon, to replicate the original score metric function  the total reward is divided by the square of the total mission period time \(T_m\) as $
\mathcal{R} = - \frac{1}{T_m^2} \sum_{t=0}^{T_m} r_t
$.

4) \textbf{Transition (\(\mathcal{T}\)):} The transition function updates the current state \(s_t\) to the next state \(s_{t+1} = (p_{t+1}, f_{t+1}, q_{t+1})\) based on the taken action \(a_t\) at time step \(t\). The agent's new position is updated to the selected task location, \(p_{t+1} = (x_{t+1}, y_{t+1}) \equiv a_t\). The mission time is updated by adding the step's elapsed time, which includes travel time between the nodes and additional recharge service time in the case of a \textit{recharging} action. It can be indicated as \(t = t_{ij} + T_R\), where \(t_{ij}\) represents the travel time between nodes \(i\) and \(j\), and \(T_R\) is the recharging time (\(T_R = 0\) for a \textit{visiting} action). The fuel level is updated as \(f_{t+1} = f_t - f_{ij}\) when a \textit{visiting} task is performed, or \(f_{t+1} = F^a\) for a \textit{recharging} task. The \textit{age period} status of the mission points is also updated as 
$\mathfrak{a}^i_{t+1} = t - t^i_{last},  \ \text{for all mission points}\ m_i \in \mathcal{M}$ and for the mission point corresponding to the selected action \(m_j \equiv a_t\), the last visit time is updated as \(t^j_{\text{last}} = t\).

\section{Problem Modelling}
\label{sec4}

In this section, we model the persistent surveillance problem using a bilevel optimization strategy as a foundational approximation. Following this, we introduce heuristics based baseline methods using a standard solver for comparison. Finally, we present our Deep Reinforcement Learning (DRL) approach as a more adaptive solution to the problem.

\subsection{Bilevel Optimization Framework}

The bilevel optimization framework or multi-echelon optimization strategies provide a structured approach for solving the UAV-UGV cooperative routing problem in persistent surveillance. These strategies are commonly employed in heterogeneous vehicle routing problems and involve multiple levels of optimization. Typically, specific sequencing strategies such as `UGV first, UAV second,' `UAV first, UGV second,' or other vehicle-prioritization schemas are used \cite{maini2015cooperation, ropero2019terra, maini2019cooperative}. The routing problem is tackled in stages: initially, in the first stage or outer level, the route for one type of vehicle (e.g., UGV or UAV) is planned based on certain assumptions about mission parameters and environmental conditions, which will serve as a reference or constraint for the subsequent optimization stage. In the second stage or inner level, the optimization of the route for the second vehicle type is carried out, considering interactions with the previously planned route of the first vehicle and adjusting for factors like task allocation, timing, and spatial coordination between the two vehicles. This hierarchical approach simplifies the complex cooperative routing problem by breaking it down into more manageable subproblems, each of which can be tackled using heuristic methods or other optimization techniques suited to the specific characteristics and capabilities of the vehicles involved. For our persistent surveillance problem modeling, we have utilized the `UGV first, UAV second' optimization strategy. In the first stage, we fix the route of the UGV, taking into consideration of its slower speed, road network constraints, and the need to serve as a mobile recharging station for the UAV. In the second stage, we construct the UAV route based on the UGV's route, optimizing for factors such as battery constraints, recharging coordination, and the necessity to revisit mission points continuously to gather real-time data.

 \subsubsection{Outer level: UGV routing}
First, we determine the UGV’s traversal direction from the starting depot along the road network $G$, which the UGV follows during the course of the persistent surveillance. This traversal also helps to determine potential refueling stops for the UAV, allowing for recharging rendezvous between the UAV and UGV. Initially, we solve a traveling salesman problem (TSP) by connecting the starting depot with the extreme end stops on the branches of the road network. The solution to the TSP provides the optimal direction for the UGV's traversal to cover the entire road network in minimum time. As illustrated in Figure \ref{UGVPlanner}\textcolor{blue}{a}, the TSP solution indicates the UGV’s travel direction. Next, we use this travel direction to identify potential locations on the road network where the UAV and UGV can meet for recharging. Let $x_0$ be the current position of the UGV and $\vec{X_g}$ be the direction of the UGV’s travel. The locations on the road network along $\vec{X_g}$, which the UGV can reach while traveling at a speed $v_g$ within the UAV's maximum flight time $T_f$,  are considered potential refueling stops provided by the UGV. The potential refuel stops $X^r$ provided by the UGV can be calculated as follows:
\begin{equation}
X^r = [ \ \{ x_i \mid \frac{|| x_i - x_0 ||}{v_g} \leqslant T_f \}, \ \forall \ x_i \in G_{\vec{X_g}}  \ ]
\label{eq:rfs}
\end{equation} 
Once the potential refuel stops offered by the UGV are identified (see Figure \ref{UGVPlanner}\textcolor{blue}{b}), the time windows for these stops are calculated. As the UAV can land on the UGV only after the UGV has reached a refueling stop, we need to determine the lower limits of the time windows as follows:
\begin{equation}
t^r_l = [ \ \{ \frac{|| x^r_i - x_0 ||}{v_g} \}, \ \forall \ {x}^{r}_i \in X^r \ ]
\end{equation}
The time windows help to synchronize the UGV and UAV at the chosen refuel stop, which is determined during UAV routing. After identifying the potential refuel stops and their time windows, this information is sent to the inner level for UAV routing. From this, we obtain rendezvous information, including the location and timing of the refueling process. After determining the refueling stop location, we connect the UGV from its current position to the chosen refuel location to obtain spatial information of the UGV route sortie as shown in Figure \ref{UGVPlanner}\textcolor{blue}{c}. Assuming a constant velocity for the UGV, we derive the temporal information of the UGV’s route sortie as well. If the UGV arrives at the refuel stop earlier than the UAV, it is assumed that the UGV will wait for the UAV.

This process is repeated iteratively each time we plan the UAV route sortie at the inner level during the persistent surveillance mission period.

 \subsubsection{Inner level: UAV routing}

 At the second stage of the optimization, we acquire the available potential refuel stop information from the UGV routing of the previous level. With the available refuel stops $X^r$ and time windows $t^r_l$, we model the UAV routing problem as an open-ended energy-constrained vehicle routing problem with time window constraints (O-EVRPTW) with the objective of
reducing the \textit{age period} of the mission points. The formulation for O-EVRPTW can be explained using graph theory. Let the available mission nodes $\mathcal{M}$ and refuel stops $X^r$ act as the vertices $V = \mathcal{M} \cup X^r $ and $ E= \{(i, j) \, \| \ i, j \, \in \, V, i \neq j \}$ denotes the edges connecting vertices $i$ and $j$. We assign a non-negative arc cost between vertices $i$ and $j$ as $t_{ij}$ (traversal time) and the decision variable as $x_{ij}$, that indicates whether a vehicle transits from $i$ to $j$; also, binary variable $y_i$ indicates if node $i$ is visited or not. The UAV commences its journey at starting point $S$, visits the mission points, and when needed, terminates its route to recharge on a UGV at any available refuel stop $X^r$, which are bounded by time-windows. The objective function, as defined in Eq. \ref{eq:1}, is to minimize the cumulative travel duration while dropping the least number of mission points and prioritizing the mission points with higher \textit{age periods} (not visited in recent times). This is achieved by imposing a penalty proportional to the square of the current \textit{age period} of the dropped mission point, $P = (\mathfrak{a}^i)^2$. This formulation ensures that the UAV prioritizes visiting mission points with a higher \textit{age period} to avoid incurring significant penalties. Eq. \ref{eq:oe} ensures that only one refuel stop is chosen where the UAV ends its route for recharging. We have established energy constraints in Eqs. \ref{eq:2} - \ref{eq:3} to make sure that the UAV never runs out of its fuel and that its fuel consumption follows the UAV’s fuel consumption rate during traversal (Eq. \ref{eq:4}). The time-window condition in Eq. \ref{eq:5} makes the UAV visits the UGV only after its arrival at any refuel stop $X^r$. Eq. \ref{eq:6} states that the cumulative arrival time at $j^{th}$ node is equal to the sum of the cumulative time at the node $i$, $t_i$ and the travel time between them $t_{ij}$. In both Eq. \ref{eq:4} \& Eq. \ref{eq:6}, we have applied Miller-Tucker Zemlin (MTZ) formulation \cite{miller1960integer} by adding large constant $L_1, L_2$ for sub-tour elimination in the UAV route. Other generic constraints for a VRP, such as flow conservation, are not shown in the paper due to space limitations. However, details can be found in our previous works \cite{mondal2023cooperative, mondal2023optimizing}.\\ \\
\text{Objective: }
\begin{equation}
 \min \quad \sum_i \sum_j t_{ij} x_{i j} + P \sum_i (1 - y_i) \quad \forall i, j \in V  \label{eq:1}
\end{equation}
\text{Major constraints:} 
\begin{equation}
 \sum_i y_i = 1  \quad \forall i \in X^r
 \label{eq:oe}
\end{equation}
\begin{equation}
f^a_i= F^a, \quad i \in X^r \label{eq:2}
\end{equation}
\begin{equation}
0 \leq f^a_i \leq F^a, \quad \forall i \in V \setminus \{S,X^r\} \label{eq:3}
\end{equation}
\begin{align}
    f^a_j &\leq f^a_i-\left(f_{i j}x_{i j}\right) \nonumber \\
    &+ L_1\left(1-x_{i j}\right), \quad \forall i,j \in V \setminus \{S,X^r\} \label{eq:4}
\end{align}
\begin{equation}
t_i  \geq t^r_{l,i}  \ , \quad \forall i \in X^r \label{eq:5}
\end{equation}
\begin{equation}
t_j \geq t_i+\left(t_{i j} x_{i j}\right)-L_2\left(1-x_{i j}\right), \quad \forall i,j \in V \label{eq:6}
\end{equation}

The UAV route sortie can be calculated by solving the O-EVRPTW formulation with any standard solver. The end coordinate and time of the sortie determine the rendezvous location $x^{\mathcal{R}}$ and rendezvous time $t^{ \mathcal{R}}$ chosen for the rendezvous (depicted in Figure \ref{UGVPlanner}\textcolor{blue}{c}). This information is then fed back to the previous level to complete the UGV sortie as described earlier. At the rendezvous position, the UAV spends a fixed recharging service time $T_R$ to complete refueling and then takes off for the next sortie.

In the persistent surveillance mission, we plan for the UGV and UAV in a receding horizon fashion, where each planning cycle corresponds to the time interval between consecutive recharging rendezvous. After each cycle, the \textit{age period} is updated based on the planned UAV and UGV routes during this interval. Once the UAV completes a recharging rendezvous, we proceed to the next planning phase to determine next sorties. This process repeats iteratively until the mission reaches the end of the overall planning horizon for the persistent surveillance task.

\subsection{Baseline Heuristics}

We establish a non-learning baseline for solving our ECUCPS problem using the bilevel optimization strategy outlined earlier. Specifically, we leverage the Google OR-Tools\texttrademark \ CP-SAT solver \cite{ORtools}, which utilizes constraint programming (CP) to solve the O-EVRPTW. To enhance the solver’s performance and avoid local optima, we implement three metaheuristics within the OR-Tools framework: 1) \textbf{Guided Local Search (GLS)}, 2) \textbf{Tabu Search (TS)}, and 3) \textbf{Simulated Annealing (SA)}. These heuristics provide approximate suboptimal solutions within significantly lesser runtime compared to traditional MILP approaches, as demonstrated in similar cooperative routing problems \cite{ramasamy2022coordinated}. This baseline approach serves as a benchmark for comparison with our proposed DRL-based methodology.

\section{Reinforcement learning framework}
\label{sec5}
This section outlines a Deep Reinforcement Learning (DRL)-based methodology for solving our UAV-UGV cooperative persistent surveillance problem. We propose an encoder-decoder-based transformer network with an RL algorithm to learn the routing policy 
$\pi_{\theta}$ for the UAV, where 
$\theta$ represents trainable parameters. Starting from the initial state $s_0$, the policy 
$\pi_{\theta}$ selects actions $a_t$ for the UAV at each timestep 
$t$, determining whether to visit a mission point or recharge based on the scenario state 
$s_t$, continuing until the terminal state 
$s_T$ is reached. The UAV's recharging action, derived from the DRL policy, determines the rendezvous instances, and the UGV's route sortie is constructed by connecting from its previous rendezvous location. The final solution of the policy network is our persistent surveillance route 
$\mathcal{T}$, consisting of a sequence of persistently visited mission points and recharging events between the UAV and UGV. This can be represented by a joint probability distribution as follows:
\begin{equation}
\mathbb{P}(\mathcal{T} ; \theta) = \prod\limits_{t=0}^{T-1} \pi_\theta(a_t | s_t) \mathbb{P}(s_{t+1} | s_t, a_t)
\end{equation}   
Here, $T$ is the timesteps till mission termination and $\mathbb{P}(s_{t+1} | s_t, a_t) = 1$, as we have chosen deterministic state transition. 

\begin{figure*}[t]
\centering
\includegraphics[scale=0.55
]{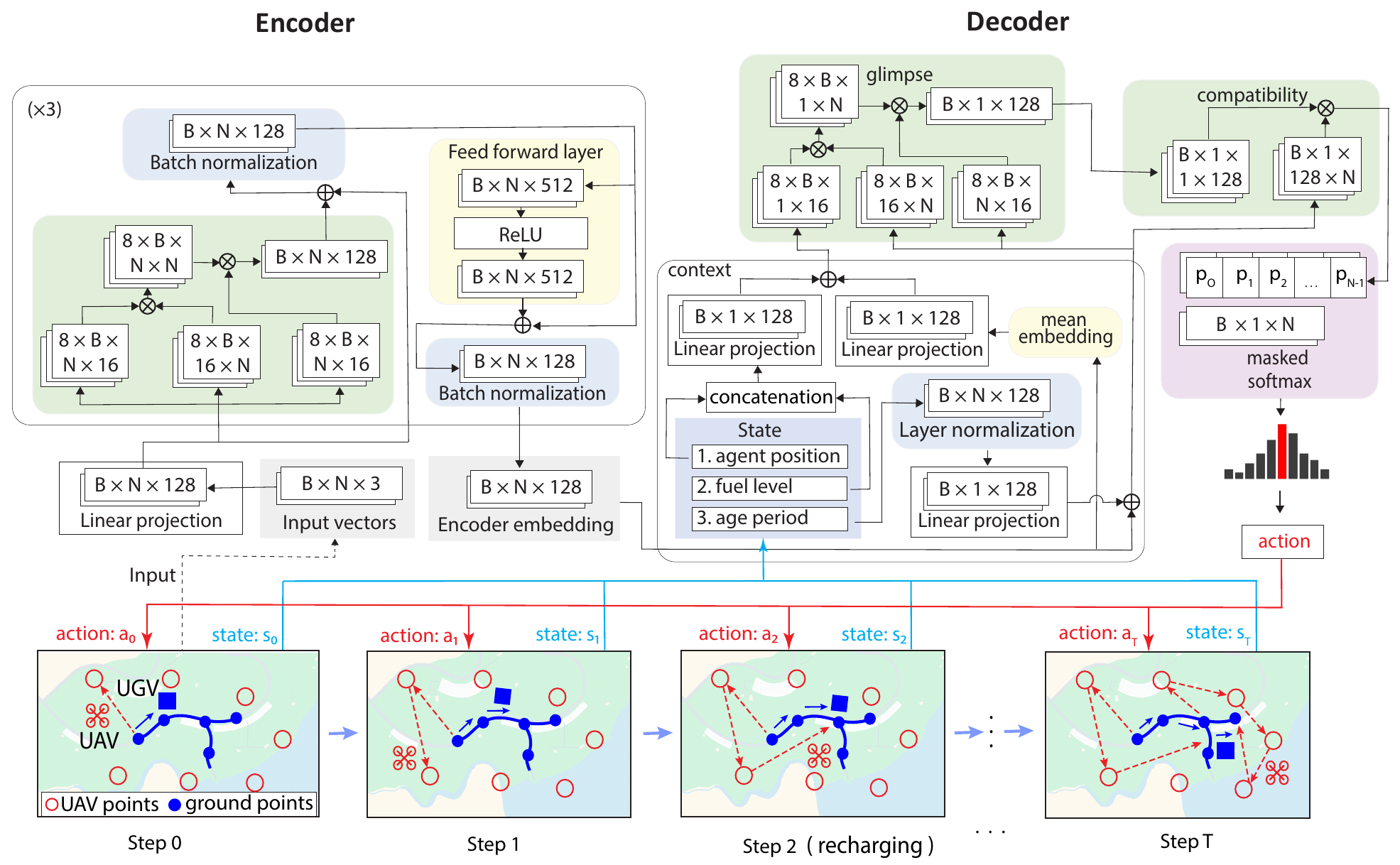}
\caption{Architecture of the proposed Transformer network. The encoder consists of three attention layers that generate input embeddings from raw data, while the decoder constructs a context vector based on the current state. The network leverages both input embeddings and the context vector, passing them through multi-head and single-head attention layers to determine the next action, sequentially forming the cooperative route for persistent surveillance, as depicted.}
\label{architecture}
\end{figure*}

\subsection{Encoder-Decoder Transformer architecture}
For the routing policy $\pi_{\theta}$, we employ an encoder-decoder transformer architecture \cite{vaswani2017attention}, which is known for its exceptional performance in natural language processing, computer vision, and vehicle routing problems due to its robust sequential data processing capabilities and efficient parallel operations \cite{xin2020step, yu2019multimodal, sun2019bert4rec}. To address the persistent surveillance problem, we adapt the transformer architecture, as depicted in Fig. \ref{architecture}. The encoder translates mission point coordinates into nuanced, high-dimensional embeddings, enhancing feature extraction. Subsequently, the decoder utilizes these node embeddings to make decisions based on \textbf{context}ual information derived from the current scenario state. This approach leverages the transformer's ability to process complex sequential data, enabling the model to effectively learn and optimize the routing policy for the UAV in the UAV-UGV cooperative persistent surveillance framework. A detailed description of the transformer architecture is provided below:
\subsubsection{Encoder} In the encoder, we have implemented multi-head attention (MHA) mechanism to achieve a higher-dimensional representation of the raw features in the problem instance. The encoder takes the 3D vector representation of mission points as input, denoted by $ X = (o_i = \{x_i, y_i, b_i\}, \forall \ m_i \in \mathcal{A}) $, where $(x_i,y_i)$ represents normalized coordinates and 
$b_i$ is a binary variable that indicates if mission point $m_i$ is eligible for recharging (ground points). Initially, each node input is linearly projected into a node embedding $h^0_i = W^0o_i + b^0$ with a dimension of
$d_h = 128$, where $W^0, b^0$ are trainable parameters for the linear projection. This initial embedding 
$h^0_i$ is then processed through $L$ identical attention layers to obtain an advanced input embedding 
$h^L_i$, facilitating a deeper understanding of relationships among the mission points. Each sublayer $ l \in L$ includes a multi-head attention (MHA) layer, a skip connection layer, a feed-forward (FF) layer, and a batch normalization (BN) layer. For the MHA layer within each sublayer $ l \in L$, three vectors, \textit{Query}, \textit{Key} and \textit{Value} are calculated from the previous layer's node embedding $h^{l-1}_i$. The dimensions of\textit{ Query}/\textit{Key}, \textit{Value} are set as
$d_q/d_k = d_v = \frac{d_h}{M} $, where $M = 8$ is the number of attention heads. For each attention head 
$j \in {1,2, ...M} $, attention scores $Z^l_j$ are calculated using \textit{Query}, \textit{Key} and \textit{Value}, which are concatenated together to produce the attention output $\text{MHA}(h^{l-1}_i)$ of that layer. The calculations are shown here:
\begin{gather}
q^l_{i,j} = h^{l-1}_iW^l_{q,j},\ k^l_{i, j} = h^{l-1}_iW^l_{k,j},\ v^l_{i, j} = h^{l-1}_iW^l_{v,j} \\
Z^l_j = \text{softmax}\left( \frac{q^l_{i,j}{k^l_{i, j}}^{T}} {\sqrt{d_k}} \right)v^l_{i, j} \\
\text{MHA}( h^{l-1}_i) = \text{Concat}(Z^l_1, Z^l_2, ..., Z^l_j) 
\end{gather}
Here, $q^l_{i,j}, k^l_{i, j} \ \text{and} \ v^l_{i, j}$ denote the \textit{Query}, \textit{Key} and \textit{Value} for head $j$, respectively, and $W^l_{q,j} \in \mathbb{R}^{d_h \times d_q}, W^l_{k,j} \in \mathbb{R}^{d_h \times d_k}$ and $W^l_{v,j} \in \mathbb{R}^{d_h \times d_v} $ are the trainable parameter matrices in the MHA layer $l$. The attention output $\text{MHA}(h^{l-1}_i)$ is then fed into a feed-forward layer (FF) with ReLU activation function to give the node embedding for that layer. Residual skip connections and Batch-Normalization (BN) are applied after both MHA and FF sublayers, as shown in the calculations below:
\begin{gather}
\hat{h}^l_i = \text{BN}(h^{l-1}_i + \text{MHA}(h^{l-1}_i)) \\
h^l_i = \text{BN}( \hat{h}^l_i + FF(\text{ReLU}(\hat{h}^l_i)))  
\end{gather}

After $L = 3$ attention layers, the final node embedding $h^L_i$ is obtained and subsequently forwarded to the decoder for further processing.

\subsubsection{Decoder}
\label{decoder_sub}
In general, during each decision-making step $t$, the decoder determines the probability of selecting each available node as an action based on the encoder’s node embedding 
$h^L_i$ and a \textbf{context} vector, which provides insights into the current scenario state. We construct the context vector $h^c_t$ by utilizing the agent’s current position embedding $h^L_{j,t}, \text{\ for \ }j \in p_t$, the agent’s current fuel level $f_t$, and the final node embedding 
$h^L_i$ obtained from the encoder. Initially, the mean graph embedding is computed from the final node embeddings to represent the global graph information by taking the average of it as shown below:
\begin{gather}
\bar{h}_t = \frac{1}{n}\sum_{i=1}^n(h^L_i) 
\end{gather}
To build the context vector, the node embedding of the agent’s current position and the current fuel level are concatenated and projected linearly. This is then combined with a linear projection of the graph embedding as shown:
\begin{gather}
h^c_t = \bar{h}_tW_g + \mbox{Cat}(h^L_{j,t}, f_t)W_c
\end{gather}
Here, $W_g, W_c$ are trainable parameters. To extract features from the current \textit{age period} status of the mission points, we utilize an \textit{age period} embedder. This embedder first performs layer normalization (LN) on the current \textit{age period} status of the mission points and subsequently projects it linearly to obtain the \textit{age period} embedding. Unlike previous works that use a fixed initial node embedding throughout the decoder's operations for computational efficiency, our approach combines the \textit{age period} embedding with the initial node embedding to create a dynamic node embedding. This dynamic embedding is then utilized in subsequent steps to make decisions. The calculations are as follows:
\begin{gather}
\hat{\mathfrak{a}}_t = \text{LN}(\mathfrak{a}_t) \\
h^a_t = \hat{\mathfrak{a}}_tW_a + h^L_i
\end{gather}
Here, $W_a$ is a trainable parameter and $\mathfrak{a}_t$ represents current \textit{age period}. The dynamic node embedding $h^a_t$ is updated at each decision-making step as the \textit{age period} status changes in the current state, capturing the essence of the mission points' visitation patterns. This helps to prioritize mission points based on the most recent information, leading to more informed and adaptive decision-making to effectively achieve the mission objective.

As the context vector and dynamic node embedding are constructed from the current state, the decoder employs a Multi-Head Attention (MHA) layer. In this layer, the context vector serves as the \textit{Query}, and the dynamic node embedding acts as the\textit{ Key }and\textit{ Value} to compute the glimpse $h^g_t$, enabling the model to simultaneously aggregate multiple aspects of the current state. The calculations are as follows:
\begin{gather}
h^g_t = \text{MHA}( h^{c}_t,\ h^a_tW^g_{k}, \ h^a_tW^g_{v}  ) 
\end{gather}
Here, $W^g_{k}, W^g_{v}$ are trainable parameter matrices. Once the glimpse 
$h^g_t$ is obtained, it is used as the \textit{Query}, and the dynamic node embedding $h^a_t$ serves as the \textit{Key} inside a single-head attention layer to compute their compatibility $h_t$. An important step here is to mask out infeasible actions based on the current scenario state. The infeasible actions include: 1) unreachable mission points at the current fuel level, 2) mission points that, if visited, would prevent the agent from reaching any refuel stop in the next step, 3) consecutive visits to the same mission point, and 4) performing recharging actions consecutively. These logical constraints are applied to determine infeasible actions, for which the compatibility is set to negative infinity. The compatibility between the \textit{Query} 
$q_t$ and \textit{Key}
$k_t$ is calculated as follows:
\begin{gather}
q_t = h^g_tW_q, k_t = h^a_tW_k \\
h_t = \begin{cases}C_p \cdot \tanh (\frac{{q_t{k_t}^T}} {\sqrt{d_q}} ) & \text { if feasible }  \\ -\infty & \text { else. } \end{cases}
\end{gather}
Here, $W_q, W_k$ are trainable matrices, and $C_p = 10$ is a clipping parameter for better exploration.  As the final step, the output probabilities for the actions are calculated using the softmax function on the compatibility: 
\begin{gather}
\pi_{\theta}(a_t | s_t) = \text{softmax}(h_t )
\end{gather}
Following the above process, the decoder sequentially determines actions to visit mission points or perform recharging until the mission is completed. We employ two decoding strategies: a greedy strategy that consistently selects the action with the highest probability at each decision-making step, and a sampling strategy that chooses actions based on their probabilities. During training, we adopt the sampling strategy to encourage better exploration. In the evaluation phase, we assess and compare the effectiveness of both strategies.

\subsection{Training method} 
The training algorithm, depicted in Algorithm \ref{algo1}, leverages the REINFORCE policy gradient method \cite{williams1992simple}. It consists of two networks: the policy network $\pi_{\theta}$, which calculates the probability distribution over actions and selects an action by sampling, and the rollout baseline network 
$\pi_{\phi}$, which has an identical structure to the policy network but selects actions greedily (choosing the action with the highest probability).

At each training iteration, routes and their corresponding rewards are calculated for a batch of instances, with expected baseline rewards for these instances are obtained from the greedy rollout in the baseline network (lines 4-13). The parameters of the policy network are updated according to the policy gradient algorithm (lines 14-16). Additionally, after each epoch, the parameters of the baseline network 
 $\phi$ are updated with those of the policy network if the baseline network performs worse than the policy network in a paired t-test (lines 17-19). 
 
 Both the policy and baseline networks are iteratively updated during the training process to yield an optimal policy by the end of training. This iterative approach ensures that the policy network consistently improves its performance through reinforcement learning while using the baseline network as a comparative measure to guide its updates effectively.

\vspace{-2mm}
\begin{figure}[htb]
\removelatexerror
\begin{algorithm}[H]
\footnotesize
\DontPrintSemicolon
\caption{Policy network training using REINFORCE algorithm} \label{algo1}  
    
\KwIn{Policy network $\pi_\theta$, Baseline network $\pi_{\phi}$, epochs $E$, Number of batches $N$, batch size $B$, episode length $T$}
\KwOut{Trained policy network $\pi_{\theta^{'}}$}

\For{epoch in $1 \dots E$}{
    Sample $N$ batches from dataset\;
    
    \For{iteration in $1 \dots N$}{
        \For{instance $b$ in $1 \dots B$}{
            Initialize $s_{0,b}$ at $t = 0$ \;
            \While{$t < T$}{
                Get action $a_{t,b} \sim \pi_\theta(a_{t,b} | s_{t,b}) $\;
                Obtain reward $r_{t,b}$ and $s_{t+1, b}$ \;
                $t = t + 1$\;
            }
            
            $\mathcal{R}_b = - \frac{1}{T_m^2} \sum_{t=0}^{T_m} r_{t,b}$\;
            Baseline reward $\mathcal{R}^{\phi}_b$ from greedy rollout with $\pi_{\phi}$\;
        }
        
        Compute gradient: \vspace{-1mm}\[\nabla_\theta J\leftarrow \frac{1}{B} \sum_{b=1}^B(\mathcal{R}_b-\mathcal{R}_b^{\phi}) \nabla_\theta \log \pi_\theta\left(s_{T, b} \mid s_{0, b}\right)\] \vspace{-4mm} \; 
        
        Update $\theta \leftarrow \theta + \alpha \nabla_\theta J$\;
    }
    
    \If{OneSidedPairedTTest$(\pi_{\theta}, \pi_{\phi}) < 0.05$}{
        $\phi \leftarrow \theta$
    }
}
\end{algorithm}
\end{figure}

\section{RESULTS}
\label{sec6}
\label{RH}

To verify the effectiveness of our proposed deep reinforcement learning framework in addressing the energy-aware UAV-UGV cooperative persistent surveillance problem, we conduct extensive computational experiments. These experiments include comparing our results with relevant baseline methodologies and performing generalization studies to evaluate the robustness of the framework. Additionally, we examine the framework's applicability for online route planning by incorporating dynamic planning and priority-driven persistent surveillance through a case study evaluation.

\subsection{Dataset details}
For training and testing, we simulate the energy-constrained UAV-UGV cooperative persistent surveillance problem over a 20 km $\times$ 20 km area, using a single UAV and a single UGV team as described in the problem statement. The UAV and UGV travel at constant speeds of 
$v_a = 10$ m/s and $v_g = 4.5$ m/s, respectively. The UAV has a fuel capacity of 
$F^a =$ 287.7 kJ and follows a fuel consumption profile defined by 
$ \mathcal{P}^a = 0.0461(v_a)^3-0.5834(v_a)^2-1.8761v_a+229.6 $, modeled after \cite{hurwitz2021mobile}, providing a maximum flight time of 25 minutes at $v_a = 10$ m/s. In this persistent surveillance problem, the UAV and UGV start from a designated depot. The UAV visits mission points both outside the road network (UAV points $\mathcal{M}_a$) and on the road network (ground points $\mathcal{M}_g$), while the UGV is restricted to the road network and can periodically recharge the UAV at any road network point to continue the mission. In the training dataset, the starting depot and UAV mission points 
$\mathcal{M}_a$ are uniformly sampled within a 4 km radius around the road network points 
$\mathcal{M}_g$. For computational efficiency, the UGV road network 
$G$ is kept fixed, but the road network points are chosen randomly from the given network. We evaluate our model on three different problem sizes: 15 UAV points with 5 ground points (\textbf{U15G5}), 30 UAV points with 10 ground points (\textbf{U30G10}), and 45 UAV points with 15 ground points (\textbf{U45G15}). The model is trained for 400 minutes of persistent surveillance mission period on U15G5 problem instances, 600 minutes on U30G10 instances, and 700 minutes on U45G15 problem instances. Longer mission periods for larger scenarios allow the agent to visit all mission points multiple times within the mission period, aiding in the learning of the optimal policy. During evaluation, the mission period is extended to allow a more comprehensive assessment of the model's performance on test scenarios. For each problem size, training is conducted on a total of 5,120,000 instances, with 256 instances per batch across 200 batches over 100 epochs. The training instances are generated on-the-fly. We use the Adam optimizer with a learning rate of 
$10^{-4}$ and a decay rate of 0.995, which is adjusted at the conclusion of each epoch as follows:
\begin{gather}
lr_{\text{decayed}} = lr \times \alpha^{n\_\text{epoch}}  
\end{gather}
 
Here, $lr$ is the current epoch's learning rate, $lr_{\text{decayed}}$ is the decayed learning rate for the next epoch, $\alpha$ is the decay rate, and $n\_\text{epoch}$ is the current epoch number. Training is completed on a server equipped with an RTX 2080 Ti GPU, with hyperparameters shared across all problem sizes. The average training time per epoch is approximately 4 minutes for U15G5, 6 minutes for U30G10, and 9 minutes for U45G15.

\subsection{Comparison evaluation}
Given the complexity and specificity of the problem, standard benchmarks are not available, and deriving an exact solution is impractical due to the problem's intractability as the number of mission points increases. Consequently, a multi-level or multi-echelon optimization strategy is a widely adopted approach for UAV-UGV cooperative routing. In persistent surveillance, as the mission points must be revisited continuously until the mission period concludes, the solution framework is often implemented in a receding horizon until a termination criterion is met. As outlined in the problem modeling (see Section \ref{sec4}), we employ a bi-level optimization strategy as the heuristic baseline method, utilizing three different metaheuristics: 1) \textbf{Guided Local Search (GLS)}, 2) \textbf{Tabu Search (TS)}, and 3) \textbf{Simulated Annealing (SA)} for comparison against the DRL policy. Additionally, we include a learning-based model, the \textbf{Attention Model (AM)} \cite{kool2018attention}, as another baseline for further comparison. During the evaluation process, our deep reinforcement learning framework employs two types of decoding strategies: a) greedy decoding, where actions with the highest probability are chosen at each decision-making step, and b) sampling decoding, where $\mathbb{N}$ trajectories are sampled to form a solution pool, from which the best solution is selected. We set the number of samples ($\mathbb{N}$) to 1024 (DRL(1024)) and 10240 (DRL(10240)) in the evaluation instances. For a fair comparison between our DRL model and the AM model, we use the same set of hyperparameters and decoding strategies during evaluation. The evaluation process is conducted using an Nvidia Quadro P2200 GPU, and the persistent surveillance mission period is extended up to 1000 minutes for the testing instances. Given the longer mission period and the increased computational time required by the heuristic approach, we evaluate 30 test instances along with all baselines. All computations are implemented in Python.
\begin{figure}[htb]
\vspace{-2mm}
\centering
\includegraphics[scale=0.40
]{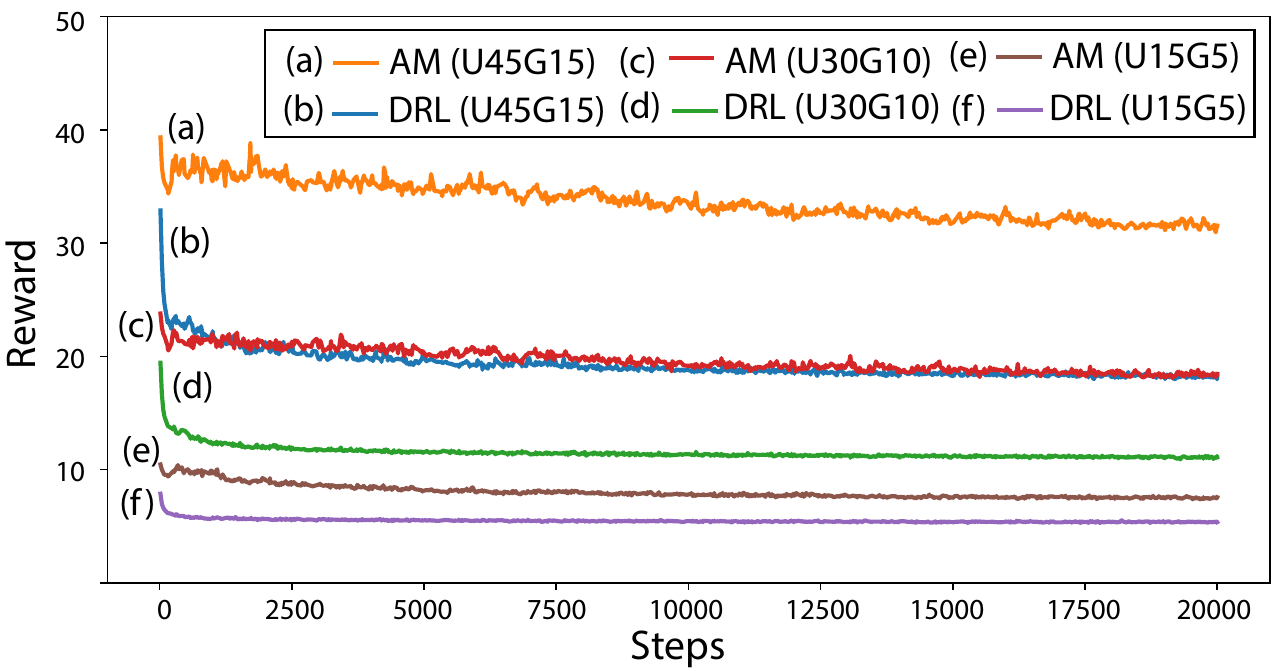}
\caption{Training reward curves across different problem sizes under DRL and AM policies. }
\label{training_curve}
\vspace{4mm}
\end{figure}

In Figure \ref{training_curve}, we present the reward training curves for our proposed DRL policy compared to the baseline AM model across three different problem sizes. Initially, both models exhibit higher rewards, which gradually converge to lower values as training progresses. Our DRL policy demonstrates superior performance by yielding lower objective function values as rewards compared to the AM model. The increase in objective function values corresponds to the increase in problem size, as more mission points contribute to a higher cumulative objective score metric. The disparity between the rewards of the DRL and AM models becomes more pronounced with larger problem sizes, with the DRL model consistently achieving lower objective scores. This indicates the enhanced capability of the DRL model in managing larger instances more effectively than the AM model. 
\begin{table}[]
\caption{Comparison evaluation of the DRL policy across problem sizes}
\renewcommand{\arraystretch}{1.1}
\centering
\small
\resizebox{\columnwidth}{!}{%
\begin{tabular}{c|ccc|ccc|ccc}
\hline
\textbf{Method} & \multicolumn{3}{c|}{\textbf{U15G5}} & \multicolumn{3}{c|}{\textbf{U30G10}} & \multicolumn{3}{c}{\textbf{U45G15}} \\
 & \textbf{Obj} & \textbf{\begin{tabular}[c]{@{}c@{}}Gap\\ (\%)\end{tabular}} & \textbf{\begin{tabular}[c]{@{}c@{}}Time\\ (sec.)\end{tabular}} & \textbf{Obj} & \textbf{\begin{tabular}[c]{@{}c@{}}Gap\\ (\%)\end{tabular}} & \textbf{\begin{tabular}[c]{@{}c@{}}Time\\ (sec)\end{tabular}} & \textbf{Obj} & \textbf{\begin{tabular}[c]{@{}c@{}}Gap\\ (\%)\end{tabular}} & \textbf{\begin{tabular}[c]{@{}c@{}}Time\\ (sec)\end{tabular}} \\ \hline
\textbf{AM(greedy)} & 7.7 & 191.3 & 2.2 & 25.9 & 296.9 & 5.6 & 42.7 & 266.8 & 6.1 \\
\textbf{DRL(greedy)} & 2.9 & 7.9 & \textbf{1.6} & 7.0 & 6.9 & \textbf{2.9} & 12.3 & 6.0 & \textbf{3.3} \\
\textbf{AM(1024)} & 3.8 & 42.3 & 6.7 & 11.5 & 76.3 & 9.8 & 22.0 & 88.9 & 9.5 \\
\textbf{DRL(1024)} & 2.7 & 0.8 & 7.3 & 6.6 & 1.1 & 10.5 & 11.8 & 1.1 & 12.8 \\
\textbf{AM(10240)} & 3.7 & 38.5 & 69.2 & 11.2 & 71.2 & 119.0 & 21.4 & 83.7 & 96.7 \\
\textbf{DRL(10240)} & \textbf{2.7} & 0.0 & 75.3 & \textbf{6.5} & 0.0 & 123.8 & \textbf{11.6} & 0.0 & 122.6 \\
\textbf{GLS} & 8.3 & 212.8 & 398.3 & 18.0 & 175.8 & 397.8 & 23.7 & 103.3 & 399.1 \\
\textbf{TS} & 3.1 & 17.0 & 366.0 & 10.5 & 60.8 & 394.2 & 20.4 & 75.0 & 399.2 \\
\textbf{SA} & 9.2 & 248.3 & 405.3 & 18.0 & 175.8 & 398.2 & 25.5 & 118.6 & 399.0 \\ \hline
\end{tabular}%
}
\label{Table1}
\end{table}

Table \ref{Table1} presents the average objective values, runtimes, and optimality gaps for problem instances across three problem sizes, comparing the methodologies. A lower objective value achieved in a shorter runtime indicates superior solution quality. The optimality gap is calculated as the difference between the objective values and the best objective function value found, using the formula:
\begin{gather}
\text{optimality gap} = \frac{\text{Obj.} - \text{Obj.}_{\text{best}}}{\text{Obj.}_{\text{best}}} \times 100\%
\end{gather}
The results show that the proposed DRL policy consistently produces lower objective values across all problem sizes compared to the baselines. Among learning-based approaches, the DRL policy outperforms the AM policy across all problem sizes with both greedy and sampling decoding strategies. The sampling decoding strategy proves more effective than the greedy strategy, as it samples more solutions and selects the best from the solution pool. The optimality gap between the AM model and the DRL policy increases with problem size, ranging from 38.5\% in smaller problems (U15G5) to 83.7\% in larger problems (U45G15), highlighting the DRL policy’s efficiency in managing larger scenarios. The DRL(10240) model achieves the best performance, producing the minimum objective scores across all problem sizes compared to the baselines. The AM model shows significant variance between its performance with greedy and sampling decoding strategies, with AM(greedy) performing the worst across all methods due to poor convergence during training. In contrast, the DRL policy shows less variation between greedy and sampling decoding, indicating good convergence of the network. Among the heuristic-based models, although the Guided Local Search (GLS) and Simulated Annealing (SA) heuristics perform worse than the learned models, the Tabu Search (TS) heuristic outperforms the baseline AM model and provides comparable performance to that of DRL policy. The optimality gap between DRL(10240) and TS heuristics ranges from 17\% in smaller problem sizes to 75\% in larger problem sizes.

In terms of computational efficiency, the DRL policy with the greedy decoding strategy yields solutions in the shortest runtime. The sampling decoding strategy takes longer due to the additional computational effort involved in solution sampling. Our DRL model generally requires slightly more runtime than the AM model, except when using greedy decoding, and the runtime increases linearly with growing problem sizes for both DRL and AM policies. The heuristic-based methods have runtimes ranging from 366 to 405 seconds, with TS heuristics being the fastest among the heuristics but still significantly slower than DRL(10240). Specifically, TS heuristics takes 5 times longer than DRL(10240) for small-scale problems, 3.5 times longer for medium-scale problems, and 3.2 times longer for large-scale problems.
Considering both solution quality and solver runtime, the DRL(greedy) model performs the best as it achieves an optimal balance between these two aspects. Animations of the routes obtained from different methods across various problem scenarios can be viewed at our website\footnote{\url{https://sites.google.com/view/ecucps}}.

\subsection{Generalization}

To analyze the generalization capability of the proposed DRL framework, we examine different testing instances by modifying two aspects of the problem scenarios: 1) increasing the problem size beyond the original training size and 2) creating problem scenarios with three different distributions of mission points to mimic diverse operational environments.
\subsubsection{Larger problem size}
In the first experiment, we assess the model’s ability to handle larger problem scenarios than its original training size by creating 30 test instances of two larger sizes. Firstly, we increase the number of mission points to a) 60 UAV points and 20 ground points (U60G20) and b) 75 UAV points and 25 ground points (U75G25). Secondly, we extend the UAV points’ spread radius to 6 km from the road network. Thirdly, the road network has also been expanded to have more ground points. We also extend the persistent surveillance mission period to 2000 minutes for these scenarios. The model’s performance is compared against the baselines on these U60G20 and U75G25 test instances, as listed in Table \ref{Table2}. The DRL(10240) policy achieves the lowest objective score among the methods. Our DRL policy demonstrates better generalization than the AM model in terms of solution quality with both greedy and sampling decoding strategies, showing an optimality gap of approximately 110\%. Overall, learning-based methods (DRL and AM) produce lower scores than heuristic methods, except for AM(greedy), which performs the worst in solution quality due to poor convergence. Among heuristics, the Tabu Search (TS) performs best with an optimality gap of approximately 120-142\%. With growing problem sizes, the total objective score increases due to the cumulative contribution from the more number of mission points. The solution runtime also increases with larger problem sizes; however, the learned models produce faster solutions than heuristic methods. Our DRL policy with the sampling decoding strategy takes longer runtime than the AM model due to additional computation; however, DRL(greedy) emerges as the fastest method, with approximately 400 times faster runtime than heuristic solutions.
\begin{table}[htb]
\caption{Comparison evaluation of the DRL policy across larger problem sizes}
\renewcommand{\arraystretch}{1.0}
\centering
\tiny
\resizebox{\columnwidth}{!}{%
\begin{tabular}{c|ccc|ccc}
\hline
\textbf{Method} & \multicolumn{3}{c|}{\textbf{U60G20}} & \multicolumn{3}{c}{\textbf{U75G25}} \\
 & \textbf{Obj} & \textbf{\begin{tabular}[c]{@{}c@{}}Gap\\ (\%)\end{tabular}} & \textbf{\begin{tabular}[c]{@{}c@{}}Time\\ (sec.)\end{tabular}} & \textbf{Obj} & \textbf{\begin{tabular}[c]{@{}c@{}}Gap\\ (\%)\end{tabular}} & \textbf{\begin{tabular}[c]{@{}c@{}}Time\\ (sec)\end{tabular}} \\ \hline
\textbf{AM(greedy)} & 60.9 & 357.7 & 10.04 & 79.9 & 326.3 & 11.26 \\
\textbf{DRL(greedy)} & 15.2 & 13.1 & \textbf{7.09} & 19.9 & 5.6 & \textbf{7.38} \\
\textbf{AM(1024)} & 28.9 & 116.4 & 19.9 & 40.4 & 115.1 & 18.62 \\
\textbf{DRL(1024)} & 13.8 & 2.2 & 28.75 & 19.0 & 1.1 & 28.1 \\
\textbf{AM(10240)} & 28.2 & 110.9 & 192.1 & 39.5 & 110.3 & 179.46 \\
\textbf{DRL(10240)} & \textbf{13.5} & 0.0 & 288.1 & \textbf{18.8} & 0.0 & 285.2 \\
\textbf{GLS} & 42.4 & 218.0 & 801.9 & 49.3 & 162.4 & 807.2 \\
\textbf{TS} & 32.3 & 142.3 & 801.7 & 41.4 & 120.6 & 807.9 \\
\textbf{SA} & 42.9 & 222.2 & 802.1 & 49.8 & 165.3 & 807.1 \\ \hline
\end{tabular}%
}
\label{Table2}
\end{table}

\begin{table*}[h]
\renewcommand{\arraystretch}{1.35}
\centering
\Huge
\caption{Comparison evaluation of the DRL policy across different mission point distributions}
\resizebox{\textwidth}{!}{%
\begin{tabular}{c|cccccclll|cccccclll|cccccclll}
\hline
\textbf{Method} & \multicolumn{9}{c|}{\textbf{Gaussian distribution}} & \multicolumn{9}{c|}{\textbf{Rayleigh distribution}} & \multicolumn{9}{c}{\textbf{Exponential distribution}} \\ \cline{2-28} 
 & \multicolumn{3}{c|}{\textbf{U15G5}} & \multicolumn{3}{c|}{\textbf{U30G10}} & \multicolumn{3}{c|}{\textbf{U45G15}} & \multicolumn{3}{c|}{\textbf{U15G5}} & \multicolumn{3}{c|}{\textbf{U30G10}} & \multicolumn{3}{c|}{\textbf{U45G15}} & \multicolumn{3}{c|}{\textbf{U15G5}} & \multicolumn{3}{c|}{\textbf{U30G10}} & \multicolumn{3}{c}{\textbf{U45G15}} \\
\textbf{} & \textbf{Obj} & \begin{tabular}[c]{@{}c@{}}\textbf{Gap}\\ (\%)\end{tabular} & \multicolumn{1}{c|}{\begin{tabular}[c]{@{}c@{}}\textbf{Time}\\ (sec)\end{tabular}} & \textbf{Obj} & \begin{tabular}[c]{@{}c@{}}\textbf{Gap}\\ (\%)\end{tabular} & \multicolumn{1}{c|}{\begin{tabular}[c]{@{}c@{}}\textbf{Time}\\ (sec)\end{tabular}} & \textbf{Obj} & \multicolumn{1}{c}{\begin{tabular}[c]{@{}c@{}}\textbf{Gap}\\ (\%)\end{tabular}} & \multicolumn{1}{c|}{\begin{tabular}[c]{@{}c@{}}\textbf{Time}\\ (sec)\end{tabular}} & \textbf{Obj} & \begin{tabular}[c]{@{}c@{}}\textbf{Gap}\\ (\%)\end{tabular} & \multicolumn{1}{c|}{\begin{tabular}[c]{@{}c@{}}\textbf{Time}\\ (sec)\end{tabular}} & \textbf{Obj} & \begin{tabular}[c]{@{}c@{}}\textbf{Gap}\\ (\%)\end{tabular} & \multicolumn{1}{c|}{\begin{tabular}[c]{@{}c@{}}\textbf{Time}\\ (sec)\end{tabular}} & \textbf{Obj} & \multicolumn{1}{c}{\begin{tabular}[c]{@{}c@{}}\textbf{Gap}\\ (\%)\end{tabular}} & \multicolumn{1}{c|}{\begin{tabular}[c]{@{}c@{}}\textbf{Time}\\ (sec)\end{tabular}} & \textbf{Obj} & \begin{tabular}[c]{@{}c@{}}\textbf{Gap}\\ (\%)\end{tabular} & \multicolumn{1}{c|}{\begin{tabular}[c]{@{}c@{}}\textbf{Time}\\ (sec)\end{tabular}} & \textbf{Obj} & \begin{tabular}[c]{@{}c@{}}\textbf{Gap}\\ (\%)\end{tabular} & \multicolumn{1}{c|}{\begin{tabular}[c]{@{}c@{}}\textbf{Time}\\ (sec)\end{tabular}} & \textbf{Obj} & \multicolumn{1}{c}{\begin{tabular}[c]{@{}c@{}}\textbf{Gap}\\ (\%)\end{tabular}} & \multicolumn{1}{c}{\begin{tabular}[c]{@{}c@{}}\textbf{Time}\\ (sec)\end{tabular}} \\ \hline
\textbf{AM(greedy)} & 8.4 & 186.0 & \multicolumn{1}{c|}{2.3} & 25.9 & 249.4 & \multicolumn{1}{c|}{4.4} & 42.4 & 215.5 & 6.5 & 10.9 & 222.8 & \multicolumn{1}{c|}{2.0} & 27.2 & 226.2 & \multicolumn{1}{c|}{5.5} & 41.6 & 172.7 & 5.7 & 8.9 & 205.2 & \multicolumn{1}{c|}{3.1} & 25.8 & 248.0 & \multicolumn{1}{c|}{3.9} & 44.3 & 249.8 & 6.5 \\
\textbf{DRL(greedy)} & 3.2 & 10.6 & \multicolumn{1}{c|}{\textbf{1.8}} & 8.1 & 9.4 & \multicolumn{1}{c|}{\textbf{2.7}} & 14.7 & 9.4 & \textbf{3.4} & 3.7 & 9.2 & \multicolumn{1}{c|}{\textbf{1.5}} & 9.2 & 10.0 & \multicolumn{1}{c|}{\textbf{2.5}} & 17.4 & 13.8 & \textbf{3.0} & 3.1 & 7.6 & \multicolumn{1}{c|}{\textbf{2.2}} & 8.1 & 8.9 & \multicolumn{1}{c|}{\textbf{2.3}} & 13.7 & 8.0 & \textbf{3.5} \\
\textbf{AM(1024)} & 4.2 & 45.2 & \multicolumn{1}{c|}{6.5} & 12.6 & 69.5 & \multicolumn{1}{c|}{9.3} & 23.7 & 76.3 & 10.2 & 5.1 & 51.9 & \multicolumn{1}{c|}{6.4} & 14.2 & 70.0 & \multicolumn{1}{c|}{9.8} & 25.8 & 69.0 & 10.7 & 4.2 & 44.8 & \multicolumn{1}{c|}{11.0} & 12.7 & 71.4 & \multicolumn{1}{c|}{9.9} & 22.8 & 79.9 & 11.7 \\
\textbf{DRL(1024)} & 3.0 & 1.4 & \multicolumn{1}{c|}{6.4} & 7.5 & 1.5 & \multicolumn{1}{c|}{10.7} & 13.7 & 1.9 & 12.3 & 3.4 & 0.9 & \multicolumn{1}{c|}{6.0} & 8.5 & 1.4 & \multicolumn{1}{c|}{9.3} & 15.5 & 1.8 & 12.3 & 2.9 & 0.7 & \multicolumn{1}{c|}{11.8} & 7.5 & 1.2 & \multicolumn{1}{c|}{10.9} & 12.8 & 1.0 & 13.7 \\
\textbf{AM(10240)} & 4.1 & 40.8 & \multicolumn{1}{c|}{67.0} & 12.2 & 64.5 & \multicolumn{1}{c|}{94.2} & 23.2 & 72.3 & 102.3 & 4.9 & 46.6 & \multicolumn{1}{c|}{68.5} & 13.7 & 63.9 & \multicolumn{1}{c|}{97.1} & 25.3 & 65.5 & 87.0 & 4.1 & 40.0 & \multicolumn{1}{c|}{103.2} & 12.2 & 64.8 & \multicolumn{1}{c|}{99.1} & 22.2 & 75.5 & 113.4 \\
\textbf{DRL(10240)} & \textbf{2.9} & 0.0 & \multicolumn{1}{c|}{82.3} & \textbf{7.4} & 0.0 & \multicolumn{1}{c|}{135.7} & \textbf{13.5} & 0.0 & 135.2 & \textbf{3.4} & 0.0 & \multicolumn{1}{c|}{63.5} & \textbf{8.3} & 0.0 & \multicolumn{1}{c|}{94.2} & \textbf{15.3} & 0.0 & 115.7 & \textbf{2.9} & 0.0 & \multicolumn{1}{c|}{74.8} & \textbf{7.4} & 0.0 & \multicolumn{1}{c|}{95.8} & \textbf{12.7} & 0.0 & 132.7 \\
\textbf{Heu GLS} & 8.5 & 190.1 & \multicolumn{1}{c|}{394.4} & 17.9 & 141.6 & \multicolumn{1}{c|}{398.1} & 27.8 & 106.5 & 399.9 & 9.2 & 173.0 & \multicolumn{1}{c|}{397.9} & 19.0 & 127.6 & \multicolumn{1}{c|}{397.8} & 28.0 & 83.7 & 399.7 & 9.0 & 210.3 & \multicolumn{1}{c|}{315.9} & 17.8 & 139.7 & \multicolumn{1}{c|}{397.6} & 25.7 & 102.8 & 399.3 \\
\textbf{Heu TS} & 3.3 & 11.3 & \multicolumn{1}{c|}{369.2} & 11.8 & 59.8 & \multicolumn{1}{c|}{395.5} & 22.2 & 64.7 & 400.2 & 3.7 & 9.5 & \multicolumn{1}{c|}{372.2} & 11.3 & 35.7 & \multicolumn{1}{c|}{393.4} & 22.1 & 44.5 & 399.8 & 3.3 & 13.8 & \multicolumn{1}{c|}{370.2} & 12.0 & 61.3 & \multicolumn{1}{c|}{395.3} & 21.5 & 69.5 & 399.3 \\
\textbf{Heu SA} & \multicolumn{1}{l}{10.2} & \multicolumn{1}{l}{250.7} & \multicolumn{1}{l|}{399.9} & \multicolumn{1}{l}{20.2} & \multicolumn{1}{l}{172.6} & \multicolumn{1}{l|}{399.7} & 29.1 & 116.4 & 399.9 & \multicolumn{1}{l}{10.8} & \multicolumn{1}{l}{220.8} & \multicolumn{1}{l|}{405.5} & \multicolumn{1}{l}{20.4} & \multicolumn{1}{l}{145.1} & \multicolumn{1}{l|}{399.8} & 30.5 & 99.7 & 400.0 & \multicolumn{1}{l}{8.1} & \multicolumn{1}{l}{178.6} & \multicolumn{1}{l|}{397.9} & \multicolumn{1}{l}{18.9} & \multicolumn{1}{l}{155.1} & \multicolumn{1}{l|}{398.1} & 28.2 & 123.0 & 399.6 \\ \hline
\end{tabular}%
}
\label{Table3}
\end{table*}

\subsubsection{Mission points distribution variations}
\begin{figure}[htp]
\centering
\includegraphics[scale=0.33
]{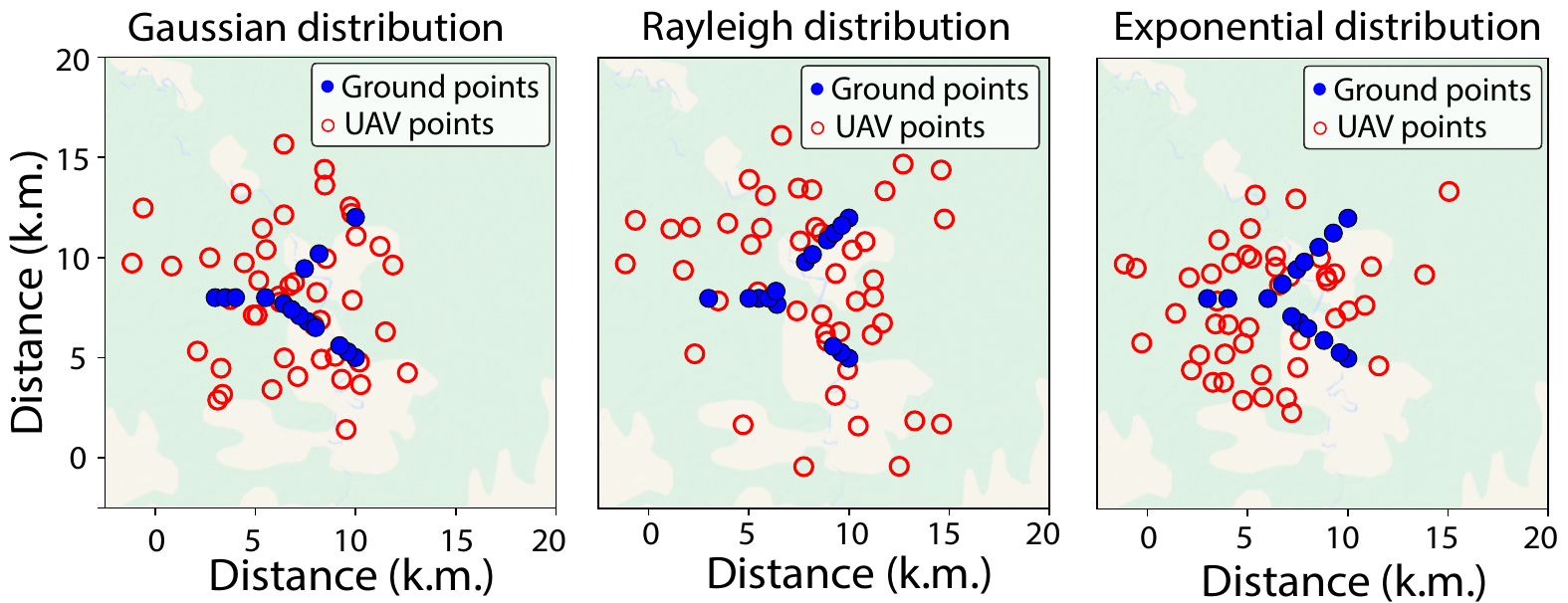}
\caption{Scenario instances from different distributions of UAV points around the road network}
\label{distributions}

\end{figure}

In our second generalization experiment, we test 30 scenarios with three different distributions of UAV mission points around the road network: a) Gaussian distribution, b) Rayleigh distribution, and c) Exponential distribution. In the Gaussian distribution, UAV points cluster around a central point, with the frequency decreasing symmetrically as the distance from the center increases. This distribution is ideal for surveillance needs concentrated around a specific location or feature, such as a building or central urban area, where events are most likely to occur. In the Rayleigh distribution, the density of UAV mission points is low near the central point and increases up to a certain radial distance, beyond which it decreases again. These scenarios are useful for situations where the area immediately around a central point does not require surveillance, but the surrounding area does, reflecting a buffer zone around a secure site. In the Exponential distribution, UAV mission points are spread such that the probability of occurrence decreases exponentially from a central point. This results in a high density of points close to the origin, with rapidly fewer points as the distance increases. This distribution is suitable for scenarios where the importance or risk decreases sharply with distance from a point of interest. Figure \ref{distributions} shows example scenarios from each type of distribution. These varied distributions help evaluate the model’s ability to generalize across different real-world scenarios, ensuring robustness and adaptability to different mission requirements. We test our proposed model along with baselines on three problem sizes (U15G5, U30G10, U45G15) of each distribution for a persistent surveillance period of 1000 minutes, and the results are represented in Table \ref{Table3}. 

Across the three distributions, our DRL(10240) policy consistently outperforms the baseline models. However, the Tabu Search (TS) heuristic model offers comparable performance to our DRL(greedy) model, with an optimality gap of 9.5-69.5\% relative to DRL(10240) across small (U15G5) to large-scale (U45G15) problems. The TS heuristic model surpasses the AM models in all problem scenarios. Conversely, the GLS and SA heuristics demonstrate poorer performance among the heuristic approaches and AM(greedy) performing the worst in terms of solution quality. Generally, the objective scores are lowest in the exponential distribution due to the centralized clustering of mission points. They increase from the Gaussian distribution to the highest in the Rayleigh distribution, attributed to the wider radial spread of mission points. The greedy decoding strategy achieves faster runtimes than the sampling decoding strategy due to reduced computation. DRL(greedy) strikes an optimal balance between time and performance, producing solutions of similar quality to TS heuristics but in significantly less time, 200 times faster in U15G5, 100 times faster in U30G10, and 65 times faster in U45G15 scenarios.

\subsection{Case study analysis: Hurricane Harvey 2017 }

\begin{figure*}[t]
\centering
\includegraphics[scale=0.5]{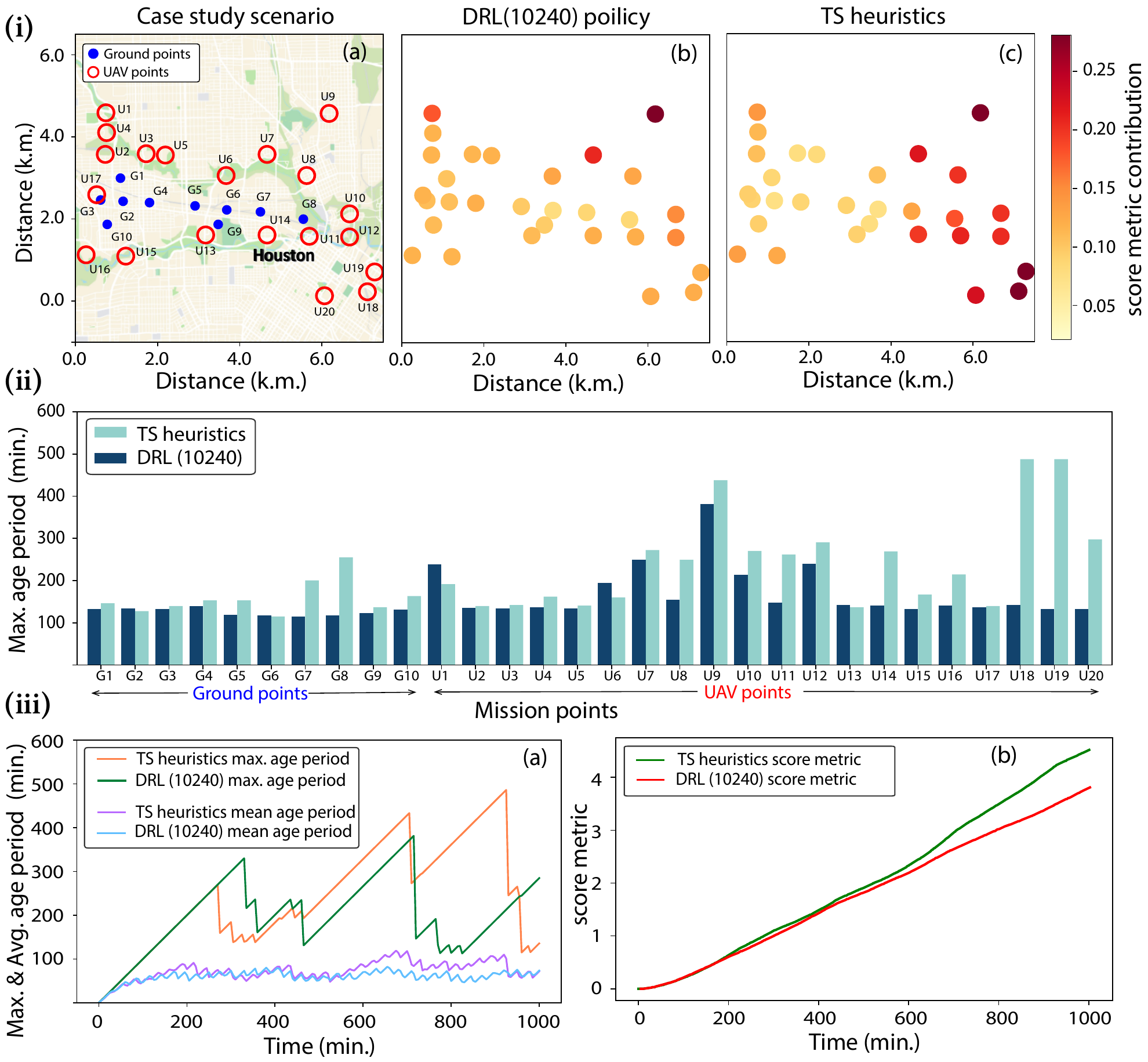} 
\caption{(Best viewed in color) Analysis of objective score metric and \textit{age period} metrics across mission points in the Hurricane Harvey 2017 case study. (i) Contribution to the objective score metric from individual mission points: 
(a) Case study scenario illustrating ground (blue) and UAV (red) mission points. (b) Heatmap showing the individual score contributions of mission points from the DRL(10240) policy. (c) Heatmap showing the individual score contributions of mission points from the TS heuristics policy. 
(ii) Maximum \textit{age period} observed at each mission point, comparing DRL(10240) and TS heuristics policies.
(iii) Comparison of \textit{age period} metrics and objective score metric over time: (a) Maximum and mean \textit{age period} across mission points during mission progression for both the DRL(10240) and TS heuristics policies. (b) Objective score metric progression throughout the mission, as obtained from both the DRL(10240) and TS heuristics policies.}

\label{case_study_fig}
\end{figure*}

\begin{figure*}[h]
\centering
\includegraphics[scale=0.425]{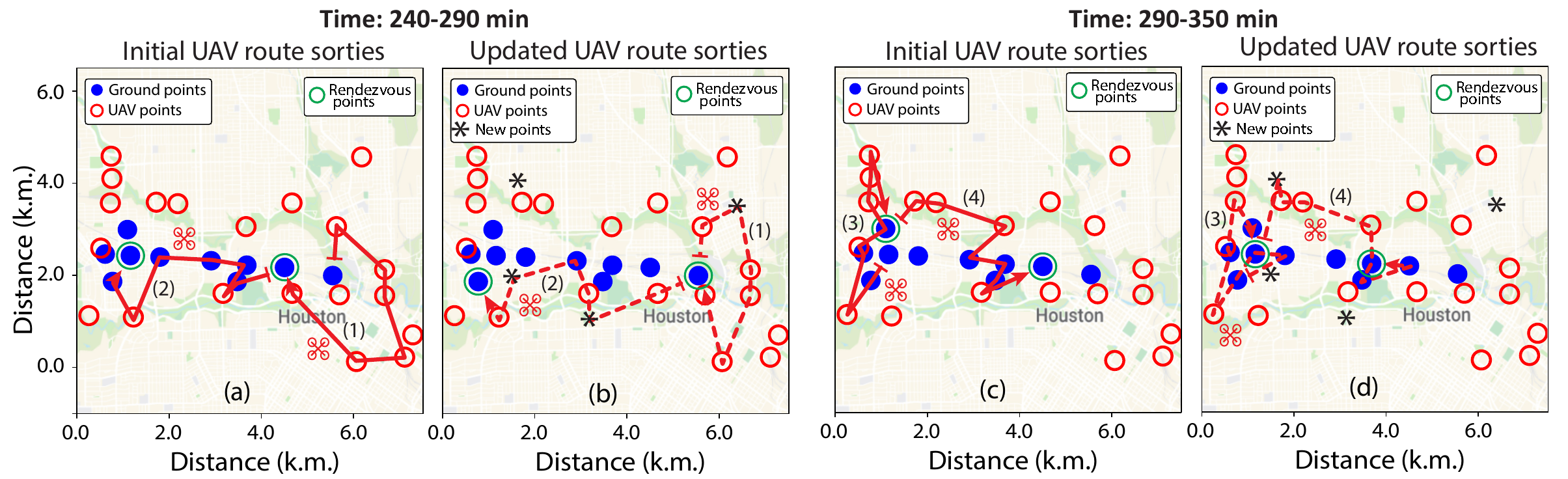} 
\caption{Dynamic route planning for newly appearing mission points (a) Initial UAV route sorties during 240–290 minutes in the case study scenario (b) Updated UAV route sorties during 240–290 minutes in response to newly added mission points (c) Initial UAV route sorties during 290–350 minutes (d) Updated UAV route sorties during 290–350 minutes in response to newly added mission points. The route animation can be found at our website.}
\label{dynamic_planning}
\end{figure*}

To evaluate the real-world applicability of our proposed framework and gain a deeper understanding of the interaction between the UAV and UGV in routing, recharging, and their impact on persistent surveillance mission performance, we apply our trained model to a case study based on the Hurricane Harvey disaster of 2017 \cite{nws_hurricane_harvey}. Hurricane Harvey, which struck in August 2017, was one of the most devastating storms in U.S. history, causing catastrophic flooding in Texas and resulting in over \$125 billion in damages. The hurricane's unprecedented rainfall, exceeding 60 inches in some areas, displaced thousands and led to lasting environmental and economic challenges, making it an ideal testbed for our case study. To construct a realistic scenario based on this disaster, we utilize the ArcGIS disaster map layer \cite{arcgis_hurricane_harvey} of Hurricane Harvey alongside the USA population density map \cite{arcgis_population_density}. The map is divided into square grids, each covering a sensor footprint of 1600 square meters, with a focus on highly populated regions around Houston. We identify 20 UAV-only mission points and 10 ground points from the road network as surveillance targets, over a 1000-minute surveillance period, particularly within flood hazard and high water extent risk zones (see Figure \ref{case_study_fig}\textcolor{blue}{ia}). Given the superior performance of the sampling decoding strategy for quality solutions, we select DRL(10240) as the DRL model and Tabu Search (TS) heuristics as the heuristic-based approach for the case study evaluation.

After conducting surveillance, we analyze the impact on individual mission points and their spatial significance by plotting the mission point-wise contributed score metric using a color map. Darker colors indicate greater contribution, reflecting longer intervals between visits or higher \textit{age periods} observed at those mission points. As shown in Figure \ref{case_study_fig}\textcolor{blue}{i}, mission points located farther from the road network and isolated from neighboring points exhibit higher \textit{score metrics}. This trend is particularly evident in the performance of the TS heuristics. The heuristic method is applied using a receding horizon approach, where each planning horizon optimizes the route sortie by solving the O-EVRPTW based on the current \textit{age period} status of mission points. However, the optimal solution at a given planning instance may not remain optimal over the entire mission period, leading to the \textit{horizon effect}. This effect occurs because the planning horizon only considers a limited timeframe, which can result in suboptimal decisions for mission points that fall outside this immediate planning window. The UAV tends to visit more closely clustered mission points within a region to achieve a higher cumulative reward from the O-EVRPTW during its planning horizon. Consequently, the far isolated points experience less frequent visits, resulting in higher \textit{age periods} and contributing significantly to the score metric. This suboptimal visitation pattern is a direct consequence of the \textit{horizon effect}, as the receding horizon approach prioritizes immediate rewards over long-term efficiency, thereby neglecting isolated mission points.
The \textit{horizon effect} is mitigated in DRL planning, as the entire cooperative route is evaluated at the end of the mission, and its performance is enhanced through a reward-based feedback mechanism. Thus, extreme mission points experience shorter \textit{age periods} under the DRL policy compared to the TS heuristics. We also calculate the maximum \textit{age period} observed at each mission point during the mission, as depicted in Figure \ref{case_study_fig}\textcolor{blue}{ii}. The TS heuristics method results in longer maximum \textit{age periods} for mission points located far from road network points. Throughout the mission, we monitor the maximum and average \textit{age periods} across all mission points, and objective score metric as shown in Figure \ref{case_study_fig}\textcolor{blue}{iii}, where the DRL policy consistently outperforms the TS heuristics. At any timestep 
$t$, the maximum and average \textit{age periods} can be defined as follows:
\begin{gather}
\text{Max. Age} = \max_{\substack{j=1,..,n}} (\mathfrak{a}^j_t) \\
\text{Avg. Age} = \frac{1}{n}\sum_{i=1}^n(\mathfrak{a}^j_t) 
\end{gather}
here, $\mathfrak{a}^j_t$ is the \textit{age period} of mission point $m_j$ at timestep $t$.

\subsection{Dynamic planning }
Our case study also investigates the potential for dynamic planning in persistent surveillance, particularly when new mission points emerge randomly during the routing process. We utilize our trained DRL policy to manage these dynamic changes for online planning. These new mission points are assumed to be located outside the road network, making them accessible only to the UAV. A critical assumption is that the UAV and UGV can exchange information exclusively during their rendezvous. Following the initial planning to establish the primary route, the trained policy updates its encoder space at each rendezvous to incorporate any new mission points, adjusting its actions as necessary. We choose the DRL(greedy) method due to its rapid execution time, which makes it highly suitable for online planning. We conduct a persistent surveillance mission lasting 1000 minutes, with the assumption that new mission points will appear at random locations within the first 200 minutes. This approach ensures that the UAV has sufficient time to visit the mission points multiple times during the remainder of the mission, enabling a comprehensive evaluation of our policy’s effectiveness.
\begin{table}[h]
\centering
\vspace{2mm}
\caption{Effect of the dynamic planning for newly added random mission points in the case study scenario}
\label{stochastics}
\begin{tabular}{ccc}
\hline
\textbf{Routes} & \textbf{score metric} & \textbf{\begin{tabular}[c]{@{}c@{}}Max. Age\\    (minutes)\end{tabular}} \\ \hline
Initial route & 3.92 & 378 \\
Modified routes & 5.51 & 523 \\ \hline
\end{tabular}
\end{table}
To assess the robustness of our model in response to the appearance of new mission points, we calculate the score metric and the maximum \textit{age period} for both the initial route and the dynamically adjusted route. We have conducted 30 trials of dynamic planning with an average of 5 newly appeared mission points, and the results detailed in Table \ref{stochastics} show little deviations in objective score metric and maximum \textit{age period} across all mission points after addition of new mission points. These findings underscore the efficacy of our DRL policy in dynamically adapting to new mission points in persistent surveillance scenarios. Figure \ref{dynamic_planning} illustrates the temporal adaptation of the UAV-UGV route obtained from the DRL policy as they adjust their initial plan to accommodate the newly emerged mission points.

\subsection {Priority driven persistent surveillance}
\begin{figure}[h]
\centering
\includegraphics[scale=0.5]{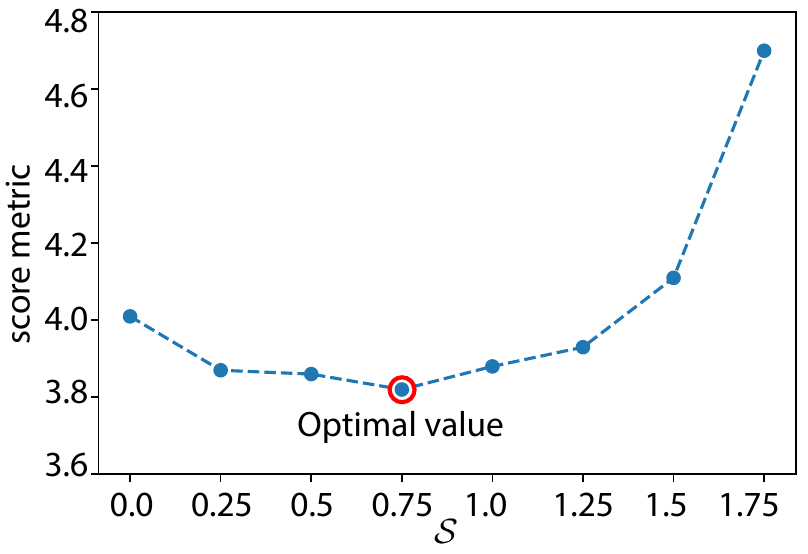} 
\caption{Variation of the objective score metric in the prioritized persistent surveillance scenario with respect to the hyperparameter $\mathcal{S}$.}
\label{s_vs_os}
\end{figure}

We extend our case study to illustrate the applicability of the DRL policy in a priority-driven persistent surveillance scenario. In this context, target mission points are assigned different weights based on their required visit frequency, which allows for prioritizing them accordingly. To demonstrate this, we modify our case study to assign higher weightage to five mission point locations that observed fewer visits (indicated by a higher maximum \textit{age period} in Figure \ref{case_study_fig}\textcolor{blue}{ii}, represented by a deeper red color in Figure \ref{case_study_fig}\textcolor{blue}{ib}) and lower weightage to another five mission points that had higher visitation rates (indicated by a lower maximum \textit{age period}, represented by a lighter yellow color in Figure \ref{case_study_fig}\textcolor{blue}{ii} and Figure \ref{case_study_fig}\textcolor{blue}{ib} respectively). The objective score metric function is also adjusted as follows:
\begin{equation}
\mathbb{S} = \frac{1}{\alpha}\sum_{k=1}^n \sum_{q=1}^p w^k\left(t_q^k-t_{q-1}^k\right)^2 \ = \frac{1}{\alpha} \sum_{k=1}^n \sum_{q=1}^p w^k(\mathfrak{a}^k_q)^2  
\end{equation}
Here, $w^k$ is the weight for mission point 
$m_k$. For high-priority mission points, we set $w^k$= 1.5, for low-priority mission points, $w^k = 0.5$, and for normal mission points, $w^k = 1$. The DRL policy decides to visit mission points based on the \textbf{context} vector that accounts for the \textit{age period} status of the mission points at each decision-making step. To adapt our DRL policy, originally trained with uniform weightage ($w^k$ = 1) for all mission points, to a priority-based scenario, we adjust the \textit{age period} status of the mission points between two timesteps by multiplying the elapsed time with an increment factor $\mathcal{F}$ and adding it to the previous \textit{age period} status, as follows:
\begin{gather}
\mathfrak{a}^k_{t+1}= \mathfrak{a}^k_t + t_{\text{elapsed}}*\mathcal{F} \\
\mathcal{F} = 1 + (w^k -1)\mathcal{S}
\end{gather}
The increment factor $\mathcal{F}$ is a function of a hyperparameter $\mathcal{S}$, which adjusts the rate of \textit{age period} increment based on the weights of the mission points. For high-priority mission points ($w^k > 1 $), 
$\mathcal{F}$ is greater than 1, leading to a more rapid increase in the \textit{age period} compared to non-weighted points. Conversely, for lower-weighted mission points ($w^k < 1 $), the \textit{age period} increment will be less. This modified \textit{age period} status allows the context vector to effectively capture the prioritization of mission points, ensuring they are visited accordingly. To optimize the hyperparameter $\mathcal{S}$, we conduct trials across a range of $\mathcal{S}$ values and select the optimal value that minimizes the objective score metric, as illustrated in Figure \ref{s_vs_os}. 

\begin{figure*}[h]
\centering
\includegraphics[scale=0.5]{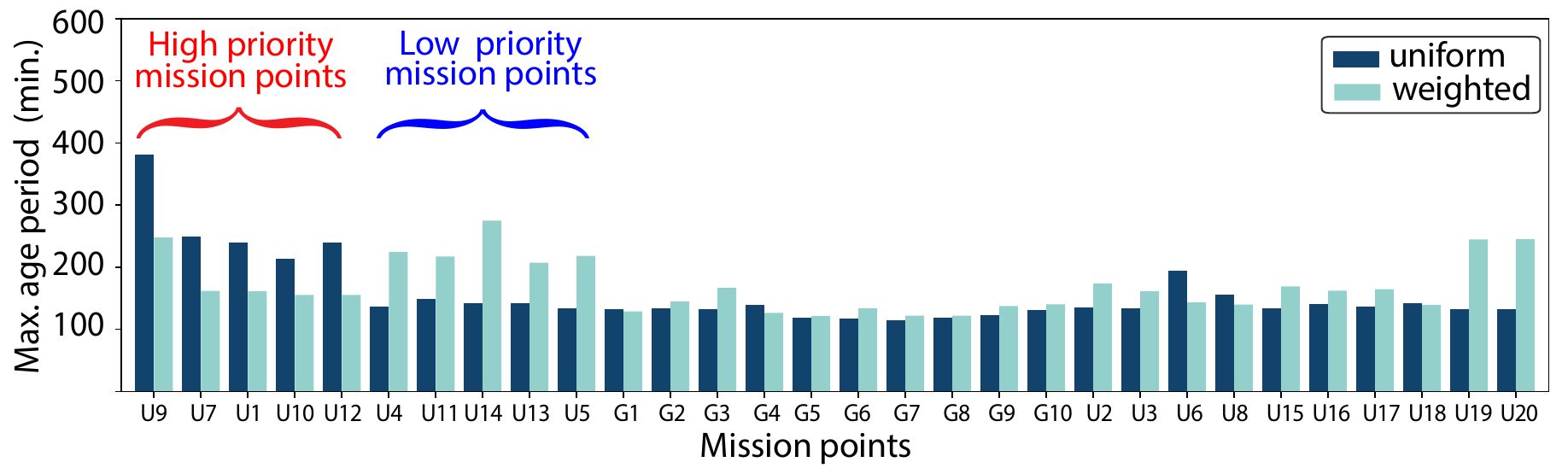} 
\caption{Maximum \textit{age periods} at individual mission points in the case study scenario, comparing uniformly weighted surveillance with priority-weighted surveillance conditions. Under priority-driven surveillance, high-priority mission points show a reduction in maximum \textit{age periods}, while low-priority mission points experience an increase in \textit{age period} values.}
\label{weighted_psr}
\end{figure*}

We have executed the routing process on the given scenario using the defined priority weights and the increment factor $\mathcal{F}$ with the optimal $\mathcal{S} = 0.75$. The maximum \textit{age period} metrics across the mission points, as derived from the routing process, is illustrated in Figure \ref{weighted_psr}. Compared to the uniformly weighted case, the results show a significant reduction in the maximum \textit{age periods} for high-priority mission points, while low-priority nodes experience a substantial increase in their maximum \textit{age periods}. This demonstrates the policy’s effectiveness in prioritizing mission points based on their assigned weights. This approach is especially valuable in disaster management scenarios when certain critical task sites require more frequent visits than less critical locations.

\section{Conclusion and future work}
\label{sec7}

In this study, we address an energy-aware UAV-UGV cooperative persistent surveillance problem to enable continuous monitoring at disaster-stricken areas, crucial for real-time observation and strategic planning in disaster management. The problem involves an energy-constrained UAV and a UGV visiting predefined mission points, with the UAV periodically recharged by the UGV acting as a mobile depot on a road network. We propose a deep reinforcement learning (DRL) framework that utilizes a transformer network with attention layers to optimize mission point visitation and UAV-UGV recharging rendezvous, ensuring prolonged surveillance. The encoder generates node embeddings from mission coordinates, while the decoder determines actions using these embeddings and contextual state. To minimize the visit-interval period (\textit{age period}) of mission points, the decoder dynamically updates the \textit{age period} embedding at each decision point, prioritizing mission points with higher \textit{age periods}. Evaluation of the proposed DRL policy on test scenarios, compared to conventional multi-stage optimization strategies (solved using heuristic approaches) and an existing learning-based model, demonstrates the following: 1) Our DRL framework outperforms baseline approaches in all problem sizes by producing lower objective score metric values in quicker runtimes. 2) It exhibits superior generalization capability, yielding better solutions in larger scenarios and across varying mission point distributions. 3) We evaluate our model on the Texas Hurricane Harvey 2017 scenario in simulation as a case study, exploring its potential for online planning by testing its adaptability to dynamic changes. 4) Our trained routing policy is successfully implemented in a priority-driven persistent surveillance scenario to address non-uniformly weighted mission points, eliminating the need for retraining.

In future work, we aim to extend the framework to support multiple UAVs and UGVs, while incorporating stochastic variations in vehicle characteristics such as speed and fuel consumption. Additionally, we plan to conduct experimental evaluations of the expanded framework using a physics-based simulation environment. Exploring stochastic planning further offers a promising direction for advancing our research and enhancing the practical application of persistent surveillance in dynamic, real-world environments.

%

\bibliographystyle{IEEEtran}	    
\bibliography{references}

\begin{thebibliography}{10}
\providecommand{\url}[1]{#1}
\csname url@samestyle\endcsname
\providecommand{\newblock}{\relax}
\providecommand{\bibinfo}[2]{#2}
\providecommand{\BIBentrySTDinterwordspacing}{\spaceskip=0pt\relax}
\providecommand{\BIBentryALTinterwordstretchfactor}{4}
\providecommand{\BIBentryALTinterwordspacing}{\spaceskip=\fontdimen2\font plus
\BIBentryALTinterwordstretchfactor\fontdimen3\font minus \fontdimen4\font\relax}
\providecommand{\BIBforeignlanguage}[2]{{%
\expandafter\ifx\csname l@#1\endcsname\relax
\typeout{** WARNING: IEEEtran.bst: No hyphenation pattern has been}%
\typeout{** loaded for the language `#1'. Using the pattern for}%
\typeout{** the default language instead.}%
\else
\language=\csname l@#1\endcsname
\fi
#2}}
\providecommand{\BIBdecl}{\relax}
\BIBdecl

\bibitem{rajan2021disaster}
J.~Rajan, S.~Shriwastav, A.~Kashyap, A.~Ratnoo, and D.~Ghose, ``Disaster management using unmanned aerial vehicles,'' in \emph{Unmanned Aerial Systems}.\hskip 1em plus 0.5em minus 0.4em\relax Elsevier, 2021, pp. 129--155.

\bibitem{erdelj2017help}
M.~Erdelj, E.~Natalizio, K.~R. Chowdhury, and I.~F. Akyildiz, ``Help from the sky: Leveraging uavs for disaster management,'' \emph{IEEE Pervasive Computing}, vol.~16, no.~1, pp. 24--32, 2017.

\bibitem{chaturvedi2019comparative}
S.~K. Chaturvedi, R.~Sekhar, S.~Banerjee, and H.~Kamal, ``Comparative review study of military and civilian unmanned aerial vehicles (uavs),'' \emph{INCAS bulletin}, vol.~11, no.~3, pp. 181--182, 2019.

\bibitem{raffetto2004unmanned}
M.~Raffetto, ``Unmanned aerial vehicle contributions to intelligence, surveillance, and reconnaissance missions for expeditionary operations,'' Ph.D. dissertation, Monterey, California. Naval Postgraduate School, 2004.

\bibitem{yamazaki2015construction}
F.~Yamazaki, T.~Matsuda, S.~Denda, and W.~Liu, ``Construction of 3d models of buildings damaged by earthquakes using uav aerial images,'' in \emph{Proceedings of the Tenth Pacific Conference on Earthquake Engineering Building an Earthquake-Resilient Pacific}, vol. 204, 2015.

\bibitem{salmoral2020guidelines}
G.~Salmoral, M.~Rivas~Casado, M.~Muthusamy, D.~Butler, P.~P. Menon, and P.~Leinster, ``Guidelines for the use of unmanned aerial systems in flood emergency response,'' \emph{Water}, vol.~12, no.~2, p. 521, 2020.

\bibitem{gupta2021uses}
K.~Gupta, S.~Bansal, and R.~Goel, ``Uses of drones in fighting covid-19 pandemic,'' in \emph{2021 10th International Conference on System Modeling \& Advancement in Research Trends (SMART)}.\hskip 1em plus 0.5em minus 0.4em\relax IEEE, 2021, pp. 651--655.

\bibitem{lin2022robust}
X.~Lin, Y.~Yaz{\i}c{\i}o{\u{g}}lu, and D.~Aksaray, ``Robust planning for persistent surveillance with energy-constrained uavs and mobile charging stations,'' \emph{IEEE Robotics and Automation Letters}, vol.~7, no.~2, pp. 4157--4164, 2022.

\bibitem{yu2019coverage}
K.~Yu, J.~M. O’Kane, and P.~Tokekar, ``Coverage of an environment using energy-constrained unmanned aerial vehicles,'' in \emph{2019 international conference on robotics and automation (ICRA)}.\hskip 1em plus 0.5em minus 0.4em\relax IEEE, 2019, pp. 3259--3265.

\bibitem{maini2015cooperation}
P.~Maini and P.~Sujit, ``On cooperation between a fuel constrained uav and a refueling ugv for large scale mapping applications,'' in \emph{2015 international conference on unmanned aircraft systems (ICUAS)}.\hskip 1em plus 0.5em minus 0.4em\relax IEEE, 2015, pp. 1370--1377.

\bibitem{wang2019vehicle}
Z.~Wang and J.-B. Sheu, ``Vehicle routing problem with drones,'' \emph{Transportation research part B: methodological}, vol. 122, pp. 350--364, 2019.

\bibitem{tang2019study}
Z.~Tang, W.-J.~v. Hoeve, and P.~Shaw, ``A study on the traveling salesman problem with a drone,'' in \emph{Integration of Constraint Programming, Artificial Intelligence, and Operations Research: 16th International Conference, CPAIOR 2019, Thessaloniki, Greece, June 4--7, 2019, Proceedings 16}.\hskip 1em plus 0.5em minus 0.4em\relax Springer, 2019, pp. 557--564.

\bibitem{sundar2016formulations}
K.~Sundar, S.~Venkatachalam, and S.~Rathinam, ``Formulations and algorithms for the multiple depot, fuel-constrained, multiple vehicle routing problem,'' in \emph{2016 American Control Conference (ACC)}.\hskip 1em plus 0.5em minus 0.4em\relax IEEE, 2016, pp. 6489--6494.

\bibitem{murray2015flying}
C.~C. Murray and A.~G. Chu, ``The flying sidekick traveling salesman problem: Optimization of drone-assisted parcel delivery,'' \emph{Transportation Research Part C: Emerging Technologies}, vol.~54, pp. 86--109, 2015.

\bibitem{cattaruzza2016vehicle}
D.~Cattaruzza, N.~Absi, and D.~Feillet, ``Vehicle routing problems with multiple trips,'' \emph{4or}, vol.~14, pp. 223--259, 2016.

\bibitem{ropero2019terra}
F.~Ropero, P.~Mu{\~n}oz, and M.~D. R-Moreno, ``Terra: A path planning algorithm for cooperative ugv--uav exploration,'' \emph{Engineering Applications of Artificial Intelligence}, vol.~78, pp. 260--272, 2019.

\bibitem{chen2019path}
M.~Chen, Y.~Chen, Z.~Chen, and Y.~Yang, ``Path planning of uav-ugv heterogeneous robot system in road network,'' in \emph{Intelligent Robotics and Applications: 12th International Conference, ICIRA 2019, Shenyang, China, August 8--11, 2019, Proceedings, Part VI 12}.\hskip 1em plus 0.5em minus 0.4em\relax Springer, 2019, pp. 497--507.

\bibitem{zhang2022cooperative}
M.~Zhang, H.~Liang, and P.~Zhou, ``Cooperative route planning for fuel-constrained ugv-uav exploration,'' in \emph{2022 IEEE International Conference on Unmanned Systems (ICUS)}.\hskip 1em plus 0.5em minus 0.4em\relax IEEE, 2022, pp. 1047--1052.

\bibitem{kaempfer2018learning}
Y.~Kaempfer and L.~Wolf, ``Learning the multiple traveling salesmen problem with permutation invariant pooling networks,'' \emph{arXiv preprint arXiv:1803.09621}, 2018.

\bibitem{wang2020learning}
Z.~Wang and M.~Gombolay, ``Learning scheduling policies for multi-robot coordination with graph attention networks,'' \emph{IEEE Robotics and Automation Letters}, vol.~5, no.~3, pp. 4509--4516, 2020.

\bibitem{sykora2020multi}
Q.~Sykora, M.~Ren, and R.~Urtasun, ``Multi-agent routing value iteration network,'' in \emph{International Conference on Machine Learning}.\hskip 1em plus 0.5em minus 0.4em\relax PMLR, 2020, pp. 9300--9310.

\bibitem{paul2022learning}
S.~Paul, P.~Ghassemi, and S.~Chowdhury, ``Learning scalable policies over graphs for multi-robot task allocation using capsule attention networks,'' in \emph{2022 International Conference on Robotics and Automation (ICRA)}.\hskip 1em plus 0.5em minus 0.4em\relax IEEE, 2022, pp. 8815--8822.

\bibitem{ramasamy2022heterogenous}
S.~Ramasamy, M.~S. Mondal, J.-P.~F. Reddinger, J.~M. Dotterweich, J.~D. Humann, M.~A. Childers, and P.~A. Bhounsule, ``Heterogenous vehicle routing: comparing parameter tuning using genetic algorithm and bayesian optimization,'' in \emph{2022 International Conference on Unmanned Aircraft Systems (ICUAS)}.\hskip 1em plus 0.5em minus 0.4em\relax IEEE, 2022, pp. 104--113.

\bibitem{mondalattention}
M.~S. Mondal, S.~Ramasamy, and P.~Bhounsule, ``An attention-aware deep reinforcement learning framework for uav-ugv collaborative route planning,'' in \emph{Proceedings of the IEEE/RSJ International Conference on Intelligent Robots and Systems (IROS)}, 2024.

\bibitem{scherer2016persistent}
J.~Scherer and B.~Rinner, ``Persistent multi-uav surveillance with energy and communication constraints,'' in \emph{2016 IEEE international conference on automation science and engineering (CASE)}.\hskip 1em plus 0.5em minus 0.4em\relax IEEE, 2016, pp. 1225--1230.

\bibitem{angun2020intelligent}
E.~Ang{\"u}n and U.~D{\"u}ndar, ``Intelligent systems for disaster management: Unmanned aerial vehicles’ surveillance problem with energy consumption constraints,'' in \emph{Intelligent and Fuzzy Techniques in Big Data Analytics and Decision Making: Proceedings of the INFUS 2019 Conference, Istanbul, Turkey, July 23-25, 2019}.\hskip 1em plus 0.5em minus 0.4em\relax Springer, 2020, pp. 1269--1278.

\bibitem{calamoneri2024management}
T.~Calamoneri, F.~Cor{\`o}, and S.~Mancini, ``Management of a post-disaster emergency scenario through unmanned aerial vehicles: Multi-depot multi-trip vehicle routing with total completion time minimization,'' \emph{Expert Systems with Applications}, vol. 251, p. 123766, 2024.

\bibitem{nigam2014multiple}
N.~Nigam, ``The multiple unmanned air vehicle persistent surveillance problem: A review,'' \emph{Machines}, vol.~2, no.~1, pp. 13--72, 2014.

\bibitem{michael2011persistent}
N.~Michael, E.~Stump, and K.~Mohta, ``Persistent surveillance with a team of mavs,'' in \emph{2011 IEEE/RSJ International Conference on Intelligent Robots and Systems}.\hskip 1em plus 0.5em minus 0.4em\relax IEEE, 2011, pp. 2708--2714.

\bibitem{hari2020optimal}
S.~K.~K. Hari, S.~Rathinam, S.~Darbha, K.~Kalyanam, S.~G. Manyam, and D.~Casbeer, ``Optimal uav route planning for persistent monitoring missions,'' \emph{IEEE Transactions on Robotics}, vol.~37, no.~2, pp. 550--566, 2020.

\bibitem{seyedi2019persistent}
S.~Seyedi, Y.~Yazicio{\u{g}}lu, and D.~Aksaray, ``Persistent surveillance with energy-constrained uavs and mobile charging stations,'' \emph{IFAC-PapersOnLine}, vol.~52, no.~20, pp. 193--198, 2019.

\bibitem{maini2019cooperative}
P.~Maini, K.~Sundar, M.~Singh, S.~Rathinam, and P.~Sujit, ``Cooperative aerial--ground vehicle route planning with fuel constraints for coverage applications,'' \emph{IEEE Transactions on Aerospace and Electronic Systems}, vol.~55, no.~6, pp. 3016--3028, 2019.

\bibitem{nigam2008persistent}
N.~Nigam and I.~Kroo, ``Persistent surveillance using multiple unmanned air vehicles,'' in \emph{2008 IEEE Aerospace Conference}.\hskip 1em plus 0.5em minus 0.4em\relax IEEE, 2008, pp. 1--14.

\bibitem{chour2022reactive}
K.~Chour, J.-P. Reddinger, J.~Dotterweich, M.~Childers, J.~Humann, S.~Rathinam, and S.~Darbha, ``A reactive energy-aware rendezvous planning approach for multi-vehicle teams,'' in \emph{2022 IEEE 18th International Conference on Automation Science and Engineering (CASE)}.\hskip 1em plus 0.5em minus 0.4em\relax IEEE, 2022, pp. 537--542.

\bibitem{asghar2023risk}
A.~B. Asghar, G.~Shi, N.~Karapetyan, J.~Humann, J.-P. Reddinger, J.~Dotterweich, and P.~Tokekar, ``Risk-aware recharging rendezvous for a collaborative team of uavs and ugvs,'' in \emph{2023 IEEE International Conference on Robotics and Automation (ICRA)}.\hskip 1em plus 0.5em minus 0.4em\relax IEEE, 2023, pp. 5544--5550.

\bibitem{vinyals2015pointer}
O.~Vinyals, M.~Fortunato, and N.~Jaitly, ``Pointer networks,'' \emph{Advances in neural information processing systems}, vol.~28, 2015.

\bibitem{bahdanau2014neural}
D.~Bahdanau, K.~Cho, and Y.~Bengio, ``Neural machine translation by jointly learning to align and translate,'' \emph{arXiv preprint arXiv:1409.0473}, 2014.

\bibitem{kool2018attention}
W.~Kool, H.~Van~Hoof, and M.~Welling, ``Attention, learn to solve routing problems!'' \emph{arXiv preprint arXiv:1803.08475}, 2018.

\bibitem{li2021deep}
J.~Li, Y.~Ma, R.~Gao, Z.~Cao, A.~Lim, W.~Song, and J.~Zhang, ``Deep reinforcement learning for solving the heterogeneous capacitated vehicle routing problem,'' \emph{IEEE Transactions on Cybernetics}, vol.~52, no.~12, pp. 13\,572--13\,585, 2021.

\bibitem{wu2021reinforcement}
G.~Wu, M.~Fan, J.~Shi, and Y.~Feng, ``Reinforcement learning based truck-and-drone coordinated delivery,'' \emph{IEEE Transactions on Artificial Intelligence}, 2021.

\bibitem{fan2022deep}
M.~Fan, Y.~Wu, T.~Liao, Z.~Cao, H.~Guo, G.~Sartoretti, and G.~Wu, ``Deep reinforcement learning for uav routing in the presence of multiple charging stations,'' \emph{IEEE Transactions on Vehicular Technology}, 2022.

\bibitem{bana2024deep}
H.~Bana, M.~Mishra, S.~Sarkar, S.~Sanjeevi, S.~PB, and K.~Sundar, ``Deep reinforcement learning-based approach for a single vehicle persistent surveillance problem with fuel constraints,'' \emph{arXiv preprint arXiv:2404.06423}, 2024.

\bibitem{miller1960integer}
C.~E. Miller, A.~W. Tucker, and R.~A. Zemlin, ``Integer programming formulation of traveling salesman problems,'' \emph{Journal of the ACM (JACM)}, vol.~7, no.~4, pp. 326--329, 1960.

\bibitem{mondal2023cooperative}
M.~S. Mondal, S.~Ramasamy, J.~D. Humann, J.-P.~F. Reddinger, J.~M. Dotterweich, M.~A. Childers, and P.~A. Bhounsule, ``Cooperative multi-agent planning framework for fuel constrained uav-ugv routing problem,'' 2023.

\bibitem{mondal2023optimizing}
M.~S. Mondal, S.~Ramasamy, J.~D. Humann, J.-P.~F. Reddinger, J.~M. Dotterweich, M.~A. Childers, and P.~Bhounsule, ``Optimizing fuel-constrained uav-ugv routes for large scale coverage: Bilevel planning in heterogeneous multi-agent systems,'' in \emph{2023 International Symposium on Multi-Robot and Multi-Agent Systems (MRS)}.\hskip 1em plus 0.5em minus 0.4em\relax IEEE, 2023, pp. 114--120.

\bibitem{ORtools}
Google, ``{Google OR-tools},'' \url{https://developers.google.com/optimization}, 2021, online; accessed Feb 2, 2021.

\bibitem{ramasamy2022coordinated}
S.~Ramasamy, J.-P.~F. Reddinger, J.~M. Dotterweich, M.~A. Childers, and P.~A. Bhounsule, ``Coordinated route planning of multiple fuel-constrained unmanned aerial systems with recharging on an unmanned ground vehicle for mission coverage,'' \emph{Journal of Intelligent \& Robotic Systems}, vol. 106, no.~1, p.~30, 2022.

\bibitem{vaswani2017attention}
A.~Vaswani, N.~Shazeer, N.~Parmar, J.~Uszkoreit, L.~Jones, A.~N. Gomez, {\L}.~Kaiser, and I.~Polosukhin, ``Attention is all you need,'' \emph{Advances in neural information processing systems}, vol.~30, 2017.

\bibitem{xin2020step}
L.~Xin, W.~Song, Z.~Cao, and J.~Zhang, ``Step-wise deep learning models for solving routing problems,'' \emph{IEEE Transactions on Industrial Informatics}, vol.~17, no.~7, pp. 4861--4871, 2020.

\bibitem{yu2019multimodal}
J.~Yu, J.~Li, Z.~Yu, and Q.~Huang, ``Multimodal transformer with multi-view visual representation for image captioning,'' \emph{IEEE transactions on circuits and systems for video technology}, vol.~30, no.~12, pp. 4467--4480, 2019.

\bibitem{sun2019bert4rec}
F.~Sun, J.~Liu, J.~Wu, C.~Pei, X.~Lin, W.~Ou, and P.~Jiang, ``Bert4rec: Sequential recommendation with bidirectional encoder representations from transformer,'' in \emph{Proceedings of the 28th ACM international conference on information and knowledge management}, 2019, pp. 1441--1450.

\bibitem{williams1992simple}
R.~J. Williams, ``Simple statistical gradient-following algorithms for connectionist reinforcement learning,'' \emph{Machine learning}, vol.~8, pp. 229--256, 1992.

\bibitem{hurwitz2021mobile}
A.~M. Hurwitz, J.~M. Dotterweich, and T.~A. Rocks, ``Mobile robot battery life estimation: battery energy use of an unmanned ground vehicle,'' in \emph{Energy Harvesting and Storage: Materials, Devices, and Applications XI}, vol. 11722.\hskip 1em plus 0.5em minus 0.4em\relax SPIE, 2021, pp. 24--40.

\bibitem{nws_hurricane_harvey}
{National Weather Service}, ``Hurricane harvey,'' \url{https://www.weather.gov/hgx/hurricaneharvey}, 2017, accessed: 2024-08-14.

\bibitem{arcgis_hurricane_harvey}
{ArcGIS}, ``Hurricane harvey disaster map,'' \url{https://www.arcgis.com/apps/View/index.html?appid=8350c2f309bb49f8865a44cb972024c2}, 2017, accessed: 2024-08-14.

\bibitem{arcgis_population_density}
Esri, ``Houston population density,'' \url{https://www.arcgis.com/apps/mapviewer/index.html?webmap=85a821d13a4f4502a85f71c4aae8bae8}, 2017, accessed: 2024-10-11.

\end{thebibliography}

\vspace{-33pt}
\begin{IEEEbiographynophoto}{Md Safwan Mondal} is a PhD student in the Department of Mechanical and Industrial Engineering at the University of Illinois Chicago. He received his BE from Jadavpur University, Kolkata. Prior to his PhD, he interned at the Indian Institute of Technology Mandi, India. His research focuses on robotics, multi-robot systems, optimization, and artificial intelligence. He has published in leading conferences such as IROS, CASE, MRS, and ICUAS, as well as in journals like JIRS, with an emphasis on multi-robot planning and optimization.
\end{IEEEbiographynophoto}
\vskip -2\baselineskip plus -1fil
\begin{IEEEbiographynophoto}{Subramanian Ramasamy } is currently pursuing a Ph.D. in the Department of Mechanical and Industrial Engineering at the University of Illinois Chicago, IL, USA. He received his B.Eng. degree from Sri Sivasubramaniya Nadar College of Engineering, Anna University, Chennai, India, in 2019. He was an Operations Research Intern at Amtrak in 2022. His research interests include operations research, deep learning, optimization, and path planning for autonomous vehicles. He has published papers in leading robotics conferences (IROS, CASE, ICUAS) and journals (JIRS) focusing on automation and route planning for unmanned vehicle systems.
\end{IEEEbiographynophoto}
\vskip -2\baselineskip plus -1fil
\begin{IEEEbiographynophoto}{Pranav Bhounsule} is an Assistant Professor in the Department of Mechanical and Industrial Engineering at the University of Illinois Chicago. He received his BE from Goa University in 2004, MTech from the Indian Institute of Technology Madras in 2006, and PhD from Cornell University in 2012. His research focuses on mathematical optimization, estimation, and control of robotic systems, with applications in unmanned systems. He has authored over 50 peer-reviewed publications and received several awards, including Best Paper in Biologically Inspired Robotics (Climbing and Walking Conference, 2012) and Best Paper Award (ASME Computers and Information in Engineering Conference, 2019).
\end{IEEEbiographynophoto}

\end{document}